\pdfoutput=1

\documentclass[11pt]{article}

\usepackage[
]{acl}
\usepackage{booktabs}
\usepackage[most]{tcolorbox}
\tcbuselibrary{listingsutf8}
\usepackage{hyperref}
\tcbuselibrary{listings, breakable}

\usepackage{times}
\usepackage{latexsym}
\usepackage{array}       %
\usepackage{pdflscape}  
\usepackage{geometry}
\usepackage{longtable}
\usepackage{enumitem}
\usepackage{float}
\usepackage{caption}
\usepackage{placeins} 
\usepackage[T1]{fontenc}
\usepackage{multirow}
\usepackage{subcaption}

\usepackage[utf8]{inputenc}
\usepackage{grffile}  
\usepackage{hyperref}
\usepackage{microtype}

\usepackage{inconsolata}
\usepackage{color}   
\usepackage{amsmath}  
\usepackage{xcolor}

\usepackage{graphicx}

\usepackage{xspace}
\usepackage{colortbl}
\usepackage{xcolor}
\usepackage{amssymb} 
\definecolor{LightRed}{rgb}{1,0.9,0.9}
\definecolor{LightGreen}{rgb}{0.9,1,0.9}

\newcommand{\methodname}{\textsc{NovelHopQA}\xspace}

\title{\methodname: Diagnosing Multi-Hop Reasoning Failures in Long Narrative Contexts}

\author{
\textbf{Abhay Gupta}\textsuperscript{1}\thanks{Lead Author} \quad
\textbf{Michael Lu}\textsuperscript{2} \\
\textbf{Kevin Zhu}\textsuperscript{1} \quad
\textbf{Sean O'Brien}\textsuperscript{1} \quad
\textbf{Vasu Sharma}\textsuperscript{3} \\
\textsuperscript{1}Algoverse AI Research \quad
\textsuperscript{2}University of California, Berkeley \quad
\textsuperscript{3}Meta FAIR Lab \\
\texttt{abhay@algoverse.us, kevin@algoverse.us}
}

\begin{document}
\maketitle

\begin{abstract}
Current large language models (LLMs) struggle to answer questions that span tens of thousands of tokens, especially when multi-hop reasoning is involved. While prior benchmarks explore long-context comprehension or multi-hop reasoning in isolation, none jointly vary context length and reasoning depth in natural narrative settings. We introduce \textbf{\methodname}, the first benchmark to evaluate k1–4 hop QA over 64k–128k-token excerpts from 83 full-length public-domain novels. A keyword-guided pipeline builds hop-separated chains grounded in coherent storylines. We evaluate seven state-of-the-art (SOTA) models and apply oracle-context filtering to ensure all questions are genuinely answerable. Human annotators validate both alignment and hop depth.We additionally present retrieval-augmented generation (RAG) evaluations to test model performance when only selected passages are provided instead of the full context. We noticed \textbf{consistent accuracy drops with increased hops and context length}, even in frontier models—revealing that sheer scale does not guarantee robust reasoning. Our failure mode analysis highlights common breakdowns, such as missed final-hop integration and long-range drift. \textbf{\methodname} offers a controlled diagnostic setting to test multi-hop reasoning at scale. All code and datasets are available at: \texttt{\href{https://novelhopqa.github.io/}{https://novelhopqa.github.io}}.
\end{abstract}

\section{Introduction}

Understanding a question whose answer is scattered across tens of thousands of tokens is still beyond today’s language models. Readers, lawyers, and historians trace clues across entire corpora, yet current NLP systems remain tuned to snippets only a few paragraphs long. When crucial evidence is buried in the middle of a long context, accuracy can plunge by more than 20 points \cite{liu2023lostmiddlelanguagemodels}. Even frontier models score below 50\% exact match on multi-document suites such as FanOutQA — where each query spans several Wikipedia pages — showing that larger context windows alone cannot solve cross-document reasoning \cite{zhu2024fanoutqamultihopmultidocumentquestion}.

Multi-hop benchmarks fall into two groups. WikiHop and HotpotQA probe two-hop reasoning over short Wikipedia passages \cite{welbl2018constructingdatasetsmultihopreading,yang2018hotpotqadatasetdiverseexplainable}. NarrativeQA, QuALITY, NovelQA, and NoCha embrace longer inputs but focus on single-hop or summary questions \cite{kočiský2017narrativeqareadingcomprehensionchallenge,pang2022qualityquestionansweringlong,wang2024novelqabenchmarkingquestionanswering,karpinska2024thousandpairsnovelchallenge}. Stress tests like MuSiQue and BABILong highlight brittleness using synthetic or stitched text \cite{trivedi2022musiquemultihopquestionssinglehop,kuratov2024babilongtestinglimitsllms}.

Standardized long-context suites — including LongBench, LEval, RULER, Marathon — show that models use a fraction of their window sizes while keeping hop depth fixed \cite{bai2024longbenchbilingualmultitaskbenchmark,an2023levalinstitutingstandardizedevaluation,hsieh2024rulerwhatsrealcontext,zhang2024marathonracerealmlong}. They do not reveal how context length interacts with reasoning depth.

Architectural advances offer partial relief. Sparse-attention models such as Longformer and BigBird reach 16–32k tokens \cite{beltagy2020longformerlongdocumenttransformer,zaheer2021bigbirdtransformerslonger}; recurrence and compression extend reach still further \cite{wu2022memorizingtransformers}; and rotary extensions break the 100 k-token barrier \cite{ding2024longropeextendingllmcontext}. Yet retrieval-augmented or attribution-guided pipelines continue to outperform context-only baselines even at 32 k+ tokens \cite{xu2024retrievalmeetslongcontext,li2024makinglongcontextlanguagemodels}. No public dataset simultaneously varies \emph{(i) hop depth} and \emph{(ii) authentic narrative context $\ge$ 64k tokens}, preventing a principled diagnosis of long-context failures.

Existing benchmarks rarely test multi-hop reasoning over long, natural context. So we ask: \textbf{can models perform multi-step reasoning across 64k–128k tokens?} We introduce \textbf{\methodname}, the first benchmark to jointly vary hop count (1–4) and narrative length, built from 83 novels with four balanced 1,000-example splits.

\vspace{0.3em}
\noindent\textbf{Contributions}
\begin{enumerate}[nosep,leftmargin=*,label=\textbf{(\arabic*)}]
    \item \textbf{Public benchmark}: 4,000 multi-hop QA examples spanning 64k–128k-token contexts.
    \item \textbf{Reproducible pipeline}: open-sourced extraction and paragraph-chaining code.
    \item \textbf{Human validation}: ten annotators confirm high alignment (\(>6.5\)/7) and hop-match accuracy (\(>94\%\)), ensuring dataset quality.
    \item \textbf{Empirical hop-depth study}: evaluations on seven SOTA models trace accuracy decay along both axes.
\end{enumerate}

Simply enlarging windows is necessary but not sufficient; true progress on long-context multi-hop reasoning demands benchmarks like \textbf{\methodname} that stress both length and depth.

\section{Related Work}

\textbf{Architectural, retrieval, and memory methods for long contexts.}  
To process longer inputs, sparse-attention and recurrence-based architectures—Longformer, BigBird, Transformer-XL, and LongRoPE—scale attention and positional encodings to tens or hundreds of thousands of tokens \cite{beltagy2020longformerlongdocumenttransformer,zaheer2021bigbirdtransformerslonger,dai2019transformerxlattentivelanguagemodels,ding2024longropeextendingllmcontext}. RAG and external-memory approaches boost performance when evidence is scattered \cite{lewis2021retrievalaugmentedgenerationknowledgeintensivenlp,wu2022memorizingtransformers}. Stress-test challenges like “Lost in the Middle” and NeedleBench highlight positional and retrieval brittleness in passages \cite{liu2023lostinmiddle,li2024needlebench}, while BABILong probes reasoning limits with synthetic million-token haystacks \cite{kuratov2024babilongtestinglimitsllms}. Although these advances surface key failure modes, they do not explore how reasoning depth interacts with very long contexts in natural prose.

\textbf{Multi-hop QA benchmarks.}  
WikiHop and HotpotQA pioneered cross-document and two-hop reasoning over short Wikipedia passages. \cite{welbl2018constructingdatasetsmultihopreading,yang2018hotpotqadatasetdiverseexplainable}. These datasets catalyzed advances in multi-hop inference but restrict inputs to at most a few thousand tokens—far from book-length scales. Subsequent compositional benchmarks such as MuSiQue introduce three-hop questions and trap-style tests \cite{trivedi2022musiquemultihopquestionssinglehop}, yet still operate on synthetic or stitched contexts rather than continuous narratives.

\textbf{Long-context QA benchmarks.}  
NarrativeQA and QuALITY probe book- or script-length inputs but mostly ask summary questions \cite{kočiský2017narrativeqareadingcomprehensionchallenge,pang2022qualityquestionansweringlong}. NoCha and NovelQA raise the ceiling to 200k tokens, with NovelQA including both single- and multi-hop questions grounded in narrative detail \cite{wang2024novelqabenchmarkingquestionanswering,karpinska2024thousandpairsnovelchallenge}. More recent datasets expand the scope further: LooGLE controls for training-data leakage while comparing short- and long-dependency reasoning over 24k+ token documents \cite{li2024looglelongcontextlanguagemodels}; LV-Eval adds five length bands up to 256k tokens and misleading facts to test robustness \cite{yuan2024lvevalbalancedlongcontextbenchmark}; and Loong focuses on multi-document QA with inputs drawn from domains like finance, law, and academia, frequently exceeding 100k tokens \cite{wang2024leavedocumentbehindbenchmarking}. FanOutQA complements these length-centric benchmarks by evaluating reasoning breadth across multiple Wikipedia pages \cite{zhu2024fanoutqamultihopmultidocumentquestion}. However, none of these benchmarks simultaneously test reasoning depth and long-context comprehension in coherent narratives—an issue that \textbf{\methodname} addresses.

\section{Dataset Construction}
\label{sec:method}

We build \textbf{\methodname}—a benchmark that probes reasoning over book‑length contexts (64k–128k tokens) with hop depths \(H \in \{1,2,3,4\}\).  
The pipeline comprises four stages: \textbf{(1)} novel selection, \textbf{(2)} anchor–keyword discovery, \textbf{(3)} paragraph chaining with incremental QA generation, and \textbf{(4)} final QA validation.  
After each hop, we regenerate the QA pair to integrate the newly appended paragraph, so the final 4‑hop item reflects four rounds of question refinement rather than a single pass at the end.

\vspace{-.1em}

\subsection{Source Corpus}
We selected 83 English novels from Project Gutenberg\footnote{\url{https://www.gutenberg.org} — All texts are in the U.S. public domain and legally permitted for research and redistribution. Our dataset annotations and processing code are released under the CC-BY-SA-4.0 license.} \cite{gutenberg}, a widely used repository of digitized books. We initially hand chose 100 diverse novels across genres and filtered this set down to 83 by removing books with fewer than 128k tokens after preprocessing. The final selection spans mystery, adventure, romance, and literary classics; includes both first- and third-person narration.

\begin{table*}[t]
\makebox[\textwidth][c]{%
  \scriptsize
  \begin{tabular}{c|ccccccc|c}
    \toprule
    \textbf{Hop}
      & \textbf{o1} & \textbf{4o} & \textbf{4o-mini}
      & \textbf{LLaMa-3.3-70B-Instruct}  
      & \textbf{Gemini 2.5 P} & \textbf{Gemini 2.0 F} & \textbf{Gemini 2.0 FL}
      & \textbf{Avg.} \\
    \midrule
    1 & 95.90 & 95.60 & 92.30 & 94.80 & \textbf{96.80} & 93.10 & 90.90 & 94.20 \\
    2 & 95.50 & 95.40 & 91.80 & 94.40 & \textbf{96.50} & 92.80 & 90.30 & 93.81 \\
    3 & 95.20 & 95.10 & 91.30 & 94.00 & \textbf{96.30} & 92.40 & 90.00 & 93.47 \\
    4 & 94.80 & 94.90 & 90.90 & 93.60 & \textbf{96.20} & 92.10 & 89.60 & 93.16 \\
    \midrule
    Avg.& 95.35 & 95.25 & 91.58 & 94.20 & \textbf{96.45} & 92.60 & 90.20 & 93.66 \\
    \bottomrule
  \end{tabular}%
}
\caption{Accuracy (\%) of each model on \textbf{\methodname} when evaluated with the original golden context.}
\label{tab:golden-context-accuracy}
\end{table*}

\begin{figure*}[t]
  \centering
  \includegraphics[width=\textwidth]{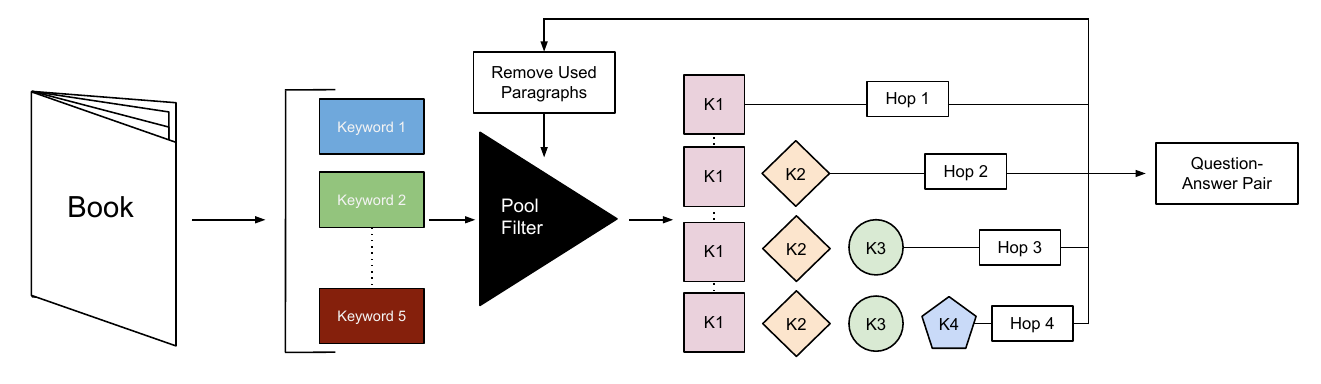}
\caption{Keyword-guided paragraph-chaining pipeline used to build \textbf{\methodname}. See Appendix~\ref{sec:example-four-hop} for a full example showing multi-hop evolution across four refinement stages.}
\label{fig:pipeline}
\end{figure*}

\begin{table*}[t]
\centering
\small
\setlength{\tabcolsep}{10pt}
\begin{tabular}{c|cccc}
\toprule
\textbf{Metric} & \textbf{$H = 1$} & \textbf{$H = 2$} & \textbf{$H = 3$} & \textbf{$H = 4$} \\
\midrule
\textbf{Alignment (1–7)} & 6.69 & 6.58 & 6.58 & 6.57 \\
\textbf{Hop Match (\%)} & 95.9 & 94.9 & 94.9 & 95.2 \\
\bottomrule
\end{tabular}
\caption{Average human validation scores across hop depths \( H \in \{1, 2, 3, 4\} \). Alignment is the mean Likert score (1–7); Hop Match is the percentage judged to require exactly \( H \) steps. See Appendix~\ref{sec:appendix-human-eval} for full table.}
\label{tab:avg_human_validation}
\end{table*}

\vspace{-.27em}

\subsection{Salient Keyword Filtering}
For each of the 83 novels, we prompt GPT‑4o‑mini \cite{openai2024gpt4technicalreport} to suggest five “anchor” keywords—characters, locations, or objects central to the plot (see Appendix~\ref{sec:appendix-prompts} for prompt).  
If any keyword appears fewer than 50 times in the text, we discard and re-sample that anchor, repeating up to seven times to ensure five high-frequency anchors.

\begin{table*}[t]
\makebox[\textwidth][c]{%
  \scriptsize
  \begin{tabular}{cc|ccccccc|c}
    \toprule
    \textbf{Context} & \textbf{Hop} 
      & \textbf{o1} & \textbf{4o} & \textbf{4o-mini}
      & \textbf{Gemini 2.5 P} & \textbf{Gemini 2.0 F} & \textbf{Gemini 2.0 FL}
      & \textbf{LLaMa-3.3-70B-Instruct} & \textbf{Avg.} \\
    \midrule
    \multirow{4}{*}{\textbf{64k}}
      & \textbf{1}
        & \textbf{92.51} & 90.12 & 75.49 & 92.34 & 87.37 & 82.53
        & 84.12 & 86.35 \\
      & \textbf{2}
        & \cellcolor{LightRed}87.66\textcolor{red}{↓4.85}
        & \cellcolor{LightRed}84.25\textcolor{red}{↓5.87}
        & \cellcolor{LightRed}74.77\textcolor{red}{↓0.72}
        & \cellcolor{LightRed}\textbf{87.84}\textcolor{red}{↓4.50}
        & \cellcolor{LightRed}77.02\textcolor{red}{↓10.35}
        & \cellcolor{LightRed}71.39\textcolor{red}{↓11.14}
        & \cellcolor{LightRed}73.88\textcolor{red}{↓10.24}
        & \cellcolor{LightRed}79.54\textcolor{red}{↓6.81} \\
      & \textbf{3}
        & \cellcolor{LightRed}84.99\textcolor{red}{↓2.67}
        & \cellcolor{LightRed}81.34\textcolor{red}{↓2.91}
        & \cellcolor{LightRed}73.14\textcolor{red}{↓1.63}
        & \cellcolor{LightRed}\textbf{85.12}\textcolor{red}{↓2.72}
        & \cellcolor{LightRed}74.25\textcolor{red}{↓2.77}
        & \cellcolor{LightRed}70.05\textcolor{red}{↓1.34}
        & \cellcolor{LightRed}71.02\textcolor{red}{↓2.86}
        & \cellcolor{LightRed}77.13\textcolor{red}{↓2.41} \\
      & \textbf{4}
        & \cellcolor{LightRed}82.15\textcolor{red}{↓2.84}
        & \cellcolor{LightRed}78.47\textcolor{red}{↓2.87}
        & \cellcolor{LightRed}68.04\textcolor{red}{↓5.10}
        & \cellcolor{LightRed}\textbf{82.45}\textcolor{red}{↓2.67}
        & \cellcolor{LightRed}71.76\textcolor{red}{↓2.49}
        & \cellcolor{LightRed}65.33\textcolor{red}{↓4.72}
        & \cellcolor{LightRed}68.11\textcolor{red}{↓2.91}
        & \cellcolor{LightRed}73.76\textcolor{red}{↓3.37} \\
    \midrule
    \multirow{4}{*}{\textbf{96k}}
      & \textbf{1}
        & \textbf{90.35} & 88.83 & 72.25 & 90.12 & 82.26 & 78.44
        & 82.04 & 83.47 \\
      & \textbf{2}
        & \cellcolor{LightRed}85.88\textcolor{red}{↓4.47}
        & \cellcolor{LightRed}82.67\textcolor{red}{↓6.16}
        & \cellcolor{LightRed}67.44\textcolor{red}{↓4.81}
        & \cellcolor{LightRed}\textbf{86.03}\textcolor{red}{↓4.09}
        & \cellcolor{LightRed}74.02\textcolor{red}{↓8.24}
        & \cellcolor{LightRed}67.04\textcolor{red}{↓11.40}
        & \cellcolor{LightRed}72.33\textcolor{red}{↓9.71}
        & \cellcolor{LightRed}76.49\textcolor{red}{↓6.98} \\
      & \textbf{3}
        & \cellcolor{LightRed}83.41\textcolor{red}{↓2.47}
        & \cellcolor{LightRed}80.41\textcolor{red}{↓2.42}
        & \cellcolor{LightRed}66.97\textcolor{red}{↓0.47}
        & \cellcolor{LightRed}\textbf{83.71}\textcolor{red}{↓2.32}
        & \cellcolor{LightRed}73.38\textcolor{red}{↓0.64}
        & \cellcolor{LightRed}66.05\textcolor{red}{↓0.99}
        & \cellcolor{LightRed}68.77\textcolor{red}{↓3.56}
        & \cellcolor{LightRed}74.67\textcolor{red}{↓1.82} \\
      & \textbf{4}
        & \cellcolor{LightRed}80.68\textcolor{red}{↓2.73}
        & \cellcolor{LightRed}76.92\textcolor{red}{↓3.91}
        & \cellcolor{LightRed}65.59\textcolor{red}{↓1.38}
        & \cellcolor{LightRed}\textbf{80.98}\textcolor{red}{↓2.73}
        & \cellcolor{LightRed}70.26\textcolor{red}{↓3.12}
        & \cellcolor{LightRed}62.81\textcolor{red}{↓3.24}
        & \cellcolor{LightRed}65.95\textcolor{red}{↓2.82}
        & \cellcolor{LightRed}71.88\textcolor{red}{↓2.79} \\
    \midrule
    \multirow{4}{*}{\textbf{128k}}
      & \textbf{1}
        & 88.76 & 86.95 & 70.03 & \textbf{89.10} & 81.77 & 75.31
        & 80.21 & 81.73 \\
      & \textbf{2}
        & \cellcolor{LightRed}84.33\textcolor{red}{↓4.43}
        & \cellcolor{LightRed}80.52\textcolor{red}{↓6.43}
        & \cellcolor{LightRed}63.95\textcolor{red}{↓6.08}
        & \cellcolor{LightRed}\textbf{84.70}\textcolor{red}{↓4.40}
        & \cellcolor{LightRed}69.13\textcolor{red}{↓12.64}
        & \cellcolor{LightRed}62.21\textcolor{red}{↓13.10}
        & \cellcolor{LightRed}69.87\textcolor{red}{↓10.34}
        & \cellcolor{LightRed}73.53\textcolor{red}{↓8.20} \\
      & \textbf{3}
        & \cellcolor{LightRed}81.92\textcolor{red}{↓2.41}
        & \cellcolor{LightRed}78.03\textcolor{red}{↓2.92}
        & \cellcolor{LightRed}62.95\textcolor{red}{↓1.00}
        & \cellcolor{LightRed}\textbf{82.20}\textcolor{red}{↓2.50}
        & \cellcolor{LightRed}68.78\textcolor{red}{↓1.35}
        & \cellcolor{LightRed}62.07\textcolor{red}{↓0.14}
        & \cellcolor{LightRed}67.92\textcolor{red}{↓1.95}
        & \cellcolor{LightRed}71.98\textcolor{red}{↓1.55} \\
      & \textbf{4}
        & \cellcolor{LightRed}\textbf{78.80}\textcolor{red}{↓3.12}
        & \cellcolor{LightRed}74.64\textcolor{red}{↓3.31}
        & \cellcolor{LightRed}61.18\textcolor{red}{↓1.77}
        & \cellcolor{LightRed}78.55\textcolor{red}{↓3.65}
        & \cellcolor{LightRed}67.32\textcolor{red}{↓1.46}
        & \cellcolor{LightRed}57.39\textcolor{red}{↓4.68}
        & \cellcolor{LightRed}64.42\textcolor{red}{↓3.50}
        & \cellcolor{LightRed}68.90\textcolor{red}{↓3.08} \\
    \bottomrule
  \end{tabular}%
}
\caption{Accuracy (\%) on \textbf{\methodname} across context lengths and hop depths, with mean performance in the last column. Red ↓ indicates drop from the previous hop; bold indicates the row-wise maximum. All cells with accuracy drops are highlighted in red. More graphs are included in Appendix~\ref{sec:appendix-breakdowns} to further visualize these trends.}
\label{tab:avg-model-performance}
\end{table*}

\subsection{Paragraph Pool Creation}
We split each novel at blank lines and discard paragraphs under 30 words.  
The remaining paragraphs form a sampling pool for context construction.

\subsection{Multi‑Hop Context Chaining \& Incremental QA Generation}
For each book and hop depth $H\in\{1,2,3,4\}$, we assemble contexts and QA pairs as follows (see Appendix~\ref{sec:appendix-prompts} for all prompts):

\begin{enumerate}[nosep,leftmargin=*]
  \item \textbf{Hop 1:}  
        Select a paragraph containing one of the book’s anchor keywords $k_1$.  
        Prompt GPT‑4o \cite{openai2024gpt4technicalreport} to generate a single‑hop QA pair $(Q_1,A_1)$ from this paragraph.
\item \textbf{Hops $h \in \{2 - H\}$:}
    \begin{enumerate}[nosep]
      \item Extract a new keyword $k_h$ from the context $C_{h-1}$ using our related‑keyword prompt. \

    \item Sample a paragraph that contains both $k_1$ and $k_h$, and append it to the growing context $C_h = C_{h-1}\,\Vert\,$new‑paragraph.
    \item Prompt GPT‑4o to re‑generate a single QA pair $(Q_h,A_h)$ over the full context $C_h$, making sure the new QA integrates evidence from all $h$ paragraphs.
    \end{enumerate}

  \item \textbf{Paragraph exclusivity:}  
        Remove selected paragraphs from the pool to prevent reuse.  
        If no matching paragraph is found after seven attempts, abort the chain and restart with a fresh anchor.
\end{enumerate}

This process “matures” each datapoint from $ (C_1,Q_1,A_1)$ through $(C_H,Q_H,A_H)$, yielding coherent multi-hop QA examples grounded in authentic narrative context. Each 64k, 96k, or 128k window is sampled from a continuous span, with all hop paragraphs required to fall within it—ensuring the QA chain reflects a cohesive narrative flow.

\subsection{Golden-Context Filtering}
\label{sec:golden_context}

To verify answerability, we evaluate all seven models on the original golden contexts used to generate each QA pair. As shown in Table~\ref{tab:golden-context-accuracy}, all models score above 90\% on average, confirming the validity of most questions. We discard any question missed by any model in the final dataset used in Section~\ref{sec:results_and_discussion}. Removal counts are reported in Appendix~\ref{sec:appendix-hopstats}.

\subsection{Irrelevant and No-Context Sanity Check}
\label{sec:methodology-fake-context}

To validate that the questions require contextual reasoning, we evaluated 800 QA pairs—100 per hop—under irrelevant and no context settings.
Removing context yields low accuracies, suggesting that the tested models are typically unable to answer correctly without contextual grounding.
This ensures the dataset reflects reasoning, not recall. Full results are in Appendix~\ref{sec:appendix-fake-context}.

\section{Human Evaluation}
Ten undergraduate validators each annotated 260 examples—40 from the 1- and 2-hop sets, and 90 from the 3- and 4-hop sets. They rated \textbf{Alignment}, measuring how well each QA pair matched its source context, and judged \textbf{Hop Match}, assessing whether the answer required exactly \( H \) reasoning steps. See Appendix~\ref{sec:appendix-human-eval} for detailed results and Appendix~\ref{sec:human-eval-form} for the evaluation form.

\section{Results and Discussion}
\label{sec:results_and_discussion}

We evaluate seven models on \textbf{\methodname} using chain-of-thought prompts: \textbf{o1} \cite{openai_o1_2024}, \textbf{Gemini 2.5 Pro} \cite{google_gemini_2_5_pro_2025}, \textbf{GPT-4o} \cite{openai2024gpt4technicalreport}, \textbf{GPT-4o-mini} \cite{openai2024gpt4technicalreport}, \textbf{LLaMA-3.3-70B-Instruct} \cite{meta2024llama33}, \textbf{Gemini 2.0 Flash}, and \textbf{Gemini 2.0 Flash Lite} \cite{google2025gemini20}. Table~\ref{tab:avg-model-performance} summarizes model accuracy across three context lengths (64k, 96k, 128k) and four hop depths (1–4).

\textbf{Impact of hop depth.}
All models show consistent performance drops as hop depth increases. On average, accuracy falls about 12 points from 1-hop to 4-hop at 64k. Even reasoning-focused models like Gemini 2.5 Pro and o1 decline steadily, with others typically dropping 14–17 points across hops.

\textbf{Impact of context length.}
Longer context windows also reduce performance, though less sharply than hop depth. Most models lose about 4–6 points from 64k to 128k on 1-hop questions. This trend is especially visible among upper-mid-tier models like GPT-4o and LLaMA, which perform well under shorter contexts but degrade more under scale.

\textbf{Model comparisons.}
Gemini 2.5 Pro and o1 consistently top each row. GPT-4o and LLaMA follow closely, both showing strong multi-hop reasoning and better robustness than smaller models. GPT-4o-mini and Flash Lite drop into the 60s under 4-hop and 128k, while Flash holds the middle but is outperformed by LLaMA at higher hops.

\textbf{Robustness at scale.}
Despite large context windows, no model maintains strong performance on the hardest tasks (4-hop at 128k), where even top models dip below 80\%. These results affirm that long-context capacity alone is not enough—multi-hop reasoning remains an open challenge. Analysis of failure modes is provided in Appendix~\ref{sec:failure_modes}, with additional RAG evaluations and breakdowns included in Appendix~\ref{sec:rag-evals}.

\section{Conclusion}  
\textbf{\methodname} is the first benchmark to vary both context length (64k–128k) and hop depth \( H \in \{1,2,3,4\} \) in long-context QA.  
Human validation confirms quality, and models show accuracy drops along both axes.  
These results highlight that \textbf{larger context windows aren't enough}—multi-hop reasoning remains a core challenge.  
We also conduct RAG evaluations~\ref{sec:rag-evals}.  
Code, data, and more details are available at: \texttt{\href{https://novelhopqa.github.io/}{https://novelhopqa.github.io}}.

\section{Limitations}
\label{sec:limitations}

\textbf{\methodname} fills a key gap in long-context, multi-hop QA, but several limitations remain:

\textbf{Genre and temporal coverage.}  
Our benchmark draws exclusively from public-domain novels available through Project Gutenberg~\cite{gutenberg}, which introduces two important limitations. First, the literary style and vocabulary reflect historical conventions of written English that may differ from contemporary usage. Second, the corpus focuses on narrative fiction while omitting other critical domains such as journalistic writing, technical documentation, and legal texts—each of which presents distinct linguistic patterns and reasoning challenges. Expanding the dataset to include modern works and non-literary genres would enhance both the diversity and practical applicability of our benchmark.

\textbf{Dialectal and domain diversity.}  
Our data largely comprises standard literary English, with few regional or archaic dialects; LLM performance on non-standard varieties may differ substantially~\cite{gupta2024aavenuedetectingllmbiases,gupta2025endivecrossdialectbenchmarkfairness}.  

\textbf{Generation and grading bias.}  
All QA pairs are generated by GPT-4o \cite{openai2024gpt4technicalreport}, and correctness is automatically graded by GPT-4.1~\cite{openai_gpt41_2024} with CoT prompts. Both steps risk inheriting model-specific patterns or blind spots. Human-authored questions and manual grading (or mixed human–machine adjudication) could reveal edge cases and reduce generator/grader artifacts.

\textbf{Evaluation metric.}  
We report accuracy as judged by GPT-4.1~\cite{openai_gpt41_2024} using CoT evaluation prompts. This approach allows for some flexibility in phrasing and considers reasoning consistency. Future evaluations could incorporate human review or rationale-based scoring for more robust assessment.

\section{Ethics Statement}
\label{sec:ethics_statement}

\textbf{Data provenance.}  
All passages are sourced from public-domain novels on Project Gutenberg~\cite{gutenberg}. No private or sensitive data is included.

\textbf{Annotator protocol.}  
Ten undergraduate validators majoring in computer science, data science, or cognitive science (aged 18+) provided informed consent and were compensated for their time. They evaluated whether each question was answerable from its context, rated alignment, and verified that the reasoning depth matched the intended hop count (Table~\ref{tab:human_validators_scores}). No additional personal data were collected.

\textbf{QA generation and grading.}  
QA pairs were generated by GPT-4o~\cite{openai2024gpt4technicalreport} and graded by GPT-4.1~\cite{openai_gpt41_2024} using CoT prompting. To validate quality, human annotators assessed whether each question aligned with its context, whether it could be answered from the provided text, and whether the reasoning depth matched the intended hop count.

\textbf{Intended use.}  
\textbf{\methodname} is provided for academic research on long-context, multi-hop reasoning. It is not intended for deployment in safety-critical or high-stakes applications without further validation.

\section*{Reproducibility Statement}

We describe our dataset construction process in Section~\ref{sec:method}, and include all prompt templates in Appendix~\ref{sec:appendix-prompts}. All model generations were obtained using publicly available APIs. Specifically, we used the Azure AI Foundry API for GPT-4o, GPT-4o-mini~\cite{openai2024gpt4technicalreport}, o1~\cite{openai_o1_2024}, and LLaMa-3.3-70B-Instruct \cite{meta2024llama33}; and the Google Vertex API for Gemini 2.0 Flash, Flash Lite~\cite{google2025gemini20}, and Gemini 2.5 Pro~\cite{google_gemini_2_5_pro_2025}. All models were queried using CoT prompts, and their outputs were graded with GPT-4.1~\cite{openai_gpt41_2024} using CoT-based evaluation prompts. We plan to release the dataset, prompts, and model outputs upon publication to support replication and further research.

\section*{Acknowledgements}

We thank Isha Sudhir, Ethan Wing, and Ernest Zhang for their helpful feedback, discussions, and support throughout the development of this work.

\bibliography{custom}

\begin{thebibliography}{38}
\providecommand{\natexlab}[1]{#1}

\bibitem[{An et~al.(2023)An, Gong, Zhong, Zhao, Li, Zhang, Kong, and Qiu}]{an2023levalinstitutingstandardizedevaluation}
Chenxin An, Shansan Gong, Ming Zhong, Xingjian Zhao, Mukai Li, Jun Zhang, Lingpeng Kong, and Xipeng Qiu. 2023.
\newblock \href {https://arxiv.org/abs/2307.11088} {L-eval: Instituting standardized evaluation for long context language models}.
\newblock \emph{Preprint}, arXiv:2307.11088.

\bibitem[{BAAI(2023)}]{bge-large-en}
BAAI. 2023.
\newblock Baai general embedding (bge-large-en).
\newblock \url{https://huggingface.co/BAAI/bge-large-en}.

\bibitem[{Bai et~al.(2024)Bai, Lv, Zhang, Lyu, Tang, Huang, Du, Liu, Zeng, Hou, Dong, Tang, and Li}]{bai2024longbenchbilingualmultitaskbenchmark}
Yushi Bai, Xin Lv, Jiajie Zhang, Hongchang Lyu, Jiankai Tang, Zhidian Huang, Zhengxiao Du, Xiao Liu, Aohan Zeng, Lei Hou, Yuxiao Dong, Jie Tang, and Juanzi Li. 2024.
\newblock \href {https://arxiv.org/abs/2308.14508} {Longbench: A bilingual, multitask benchmark for long context understanding}.
\newblock \emph{Preprint}, arXiv:2308.14508.

\bibitem[{Beltagy et~al.(2020)Beltagy, Peters, and Cohan}]{beltagy2020longformerlongdocumenttransformer}
Iz~Beltagy, Matthew~E. Peters, and Arman Cohan. 2020.
\newblock \href {https://arxiv.org/abs/2004.05150} {Longformer: The long-document transformer}.
\newblock \emph{Preprint}, arXiv:2004.05150.

\bibitem[{Dai et~al.(2019)Dai, Yang, Yang, Carbonell, Le, and Salakhutdinov}]{dai2019transformerxlattentivelanguagemodels}
Zihang Dai, Zhilin Yang, Yiming Yang, Jaime Carbonell, Quoc~V. Le, and Ruslan Salakhutdinov. 2019.
\newblock \href {https://arxiv.org/abs/1901.02860} {Transformer-xl: Attentive language models beyond a fixed-length context}.
\newblock \emph{Preprint}, arXiv:1901.02860.

\bibitem[{DeepMind(2025{\natexlab{a}})}]{google2025gemini20}
Google DeepMind. 2025{\natexlab{a}}.
\newblock \href {https://cloud.google.com/vertex-ai/generative-ai/docs/models/gemini/2-0-flash} {Gemini 2.0 flash and flash lite}.
\newblock Online documentation.
\newblock Google Cloud, Vertex AI, and Google AI Studio documentation.

\bibitem[{DeepMind(2025{\natexlab{b}})}]{google_gemini_2_5_pro_2025}
Google DeepMind. 2025{\natexlab{b}}.
\newblock Gemini model and thinking updates: March 2025.
\newblock \url{https://blog.google/technology/google-deepmind/gemini-model-thinking-updates-march-2025/}.
\newblock Accessed: 2025-05-16.

\bibitem[{Ding et~al.(2024)Ding, Zhang, Zhang, Xu, Shang, Xu, Yang, and Yang}]{ding2024longropeextendingllmcontext}
Yiran Ding, Li~Lyna Zhang, Chengruidong Zhang, Yuanyuan Xu, Ning Shang, Jiahang Xu, Fan Yang, and Mao Yang. 2024.
\newblock \href {https://arxiv.org/abs/2402.13753} {Longrope: Extending llm context window beyond 2 million tokens}.
\newblock \emph{Preprint}, arXiv:2402.13753.

\bibitem[{Gupta et~al.(2025)Gupta, Cheung, Meng, Sayyed, Liao, Zhu, and O'Brien}]{gupta2025endivecrossdialectbenchmarkfairness}
Abhay Gupta, Jacob Cheung, Philip Meng, Shayan Sayyed, Austen Liao, Kevin Zhu, and Sean O'Brien. 2025.
\newblock \href {https://arxiv.org/abs/2504.07100} {Endive: A cross-dialect benchmark for fairness and performance in large language models}.
\newblock \emph{Preprint}, arXiv:2504.07100.

\bibitem[{Gupta et~al.(2024)Gupta, Meng, Yurtseven, O'Brien, and Zhu}]{gupta2024aavenuedetectingllmbiases}
Abhay Gupta, Philip Meng, Ece Yurtseven, Sean O'Brien, and Kevin Zhu. 2024.
\newblock \href {https://arxiv.org/abs/2408.14845} {Aavenue: Detecting llm biases on nlu tasks in aave via a novel benchmark}.
\newblock \emph{Preprint}, arXiv:2408.14845.

\bibitem[{Gutenberg(2025)}]{gutenberg}
Project Gutenberg. 2025.
\newblock \href {https://www.gutenberg.org/} {Project gutenberg}.
\newblock Accessed: 2025-04-17.

\bibitem[{Hsieh et~al.(2024)Hsieh, Sun, Kriman, Acharya, Rekesh, Jia, Zhang, and Ginsburg}]{hsieh2024rulerwhatsrealcontext}
Cheng-Ping Hsieh, Simeng Sun, Samuel Kriman, Shantanu Acharya, Dima Rekesh, Fei Jia, Yang Zhang, and Boris Ginsburg. 2024.
\newblock \href {https://arxiv.org/abs/2404.06654} {Ruler: What's the real context size of your long-context language models?}
\newblock \emph{Preprint}, arXiv:2404.06654.

\bibitem[{Karpinska et~al.(2024)Karpinska, Thai, Lo, Goyal, and Iyyer}]{karpinska2024thousandpairsnovelchallenge}
Marzena Karpinska, Katherine Thai, Kyle Lo, Tanya Goyal, and Mohit Iyyer. 2024.
\newblock \href {https://arxiv.org/abs/2406.16264} {One thousand and one pairs: A "novel" challenge for long-context language models}.
\newblock \emph{Preprint}, arXiv:2406.16264.

\bibitem[{Kočiský et~al.(2017)Kočiský, Schwarz, Blunsom, Dyer, Hermann, Melis, and Grefenstette}]{kočiský2017narrativeqareadingcomprehensionchallenge}
Tomáš Kočiský, Jonathan Schwarz, Phil Blunsom, Chris Dyer, Karl~Moritz Hermann, Gábor Melis, and Edward Grefenstette. 2017.
\newblock \href {https://arxiv.org/abs/1712.07040} {The narrativeqa reading comprehension challenge}.
\newblock \emph{Preprint}, arXiv:1712.07040.

\bibitem[{Kuratov et~al.(2024)Kuratov, Bulatov, Anokhin, Rodkin, Sorokin, Sorokin, and Burtsev}]{kuratov2024babilongtestinglimitsllms}
Yuri Kuratov, Aydar Bulatov, Petr Anokhin, Ivan Rodkin, Dmitry Sorokin, Artyom Sorokin, and Mikhail Burtsev. 2024.
\newblock \href {https://arxiv.org/abs/2406.10149} {Babilong: Testing the limits of llms with long context reasoning-in-a-haystack}.
\newblock \emph{Preprint}, arXiv:2406.10149.

\bibitem[{Lewis et~al.(2021)Lewis, Perez, Piktus, Petroni, Karpukhin, Goyal, Küttler, Lewis, tau Yih, Rocktäschel, Riedel, and Kiela}]{lewis2021retrievalaugmentedgenerationknowledgeintensivenlp}
Patrick Lewis, Ethan Perez, Aleksandra Piktus, Fabio Petroni, Vladimir Karpukhin, Naman Goyal, Heinrich Küttler, Mike Lewis, Wen tau Yih, Tim Rocktäschel, Sebastian Riedel, and Douwe Kiela. 2021.
\newblock \href {https://arxiv.org/abs/2005.11401} {Retrieval-augmented generation for knowledge-intensive nlp tasks}.
\newblock \emph{Preprint}, arXiv:2005.11401.

\bibitem[{Li et~al.(2024{\natexlab{a}})Li, Wang, Zheng, and Zhang}]{li2024looglelongcontextlanguagemodels}
Jiaqi Li, Mengmeng Wang, Zilong Zheng, and Muhan Zhang. 2024{\natexlab{a}}.
\newblock \href {https://arxiv.org/abs/2311.04939} {Loogle: Can long-context language models understand long contexts?}
\newblock \emph{Preprint}, arXiv:2311.04939.

\bibitem[{Li et~al.(2024{\natexlab{b}})Li, , Zhang, Liu, and Chen}]{li2024needlebench}
Mo~Li, , Songyang Zhang, Yunxin Liu, and Kai Chen. 2024{\natexlab{b}}.
\newblock \href {https://arxiv.org/abs/2407.11963} {Needlebench: Can llms do retrieval and reasoning in 1 million context window?}
\newblock \emph{Preprint}, arXiv:2407.11963.

\bibitem[{Li et~al.(2024{\natexlab{c}})Li, Liang, Lyu, and Wang}]{li2024makinglongcontextlanguagemodels}
Yanyang Li, Shuo Liang, Michael~R. Lyu, and Liwei Wang. 2024{\natexlab{c}}.
\newblock \href {https://arxiv.org/abs/2408.03246} {Making long-context language models better multi-hop reasoners}.
\newblock \emph{Preprint}, arXiv:2408.03246.

\bibitem[{Liu et~al.(2023{\natexlab{a}})Liu, Lin, Hewitt, Paranjape, Bevilacqua, Petroni, and Liang}]{liu2023lostmiddlelanguagemodels}
Nelson~F. Liu, Kevin Lin, John Hewitt, Ashwin Paranjape, Michele Bevilacqua, Fabio Petroni, and Percy Liang. 2023{\natexlab{a}}.
\newblock \href {https://arxiv.org/abs/2307.03172} {Lost in the middle: How language models use long contexts}.
\newblock \emph{Preprint}, arXiv:2307.03172.

\bibitem[{Liu et~al.(2023{\natexlab{b}})Liu, Lin, Hewitt, Paranjape, Bevilacqua, Petroni, and Liang}]{liu2023lostinmiddle}
Nelson~F. Liu, Kevin Lin, John Hewitt, Ashwin Paranjape, Michele Bevilacqua, Fabio Petroni, and Percy Liang. 2023{\natexlab{b}}.
\newblock \href {https://arxiv.org/abs/2307.03172} {Lost in the middle: How language models use long contexts}.
\newblock \emph{Preprint}, arXiv:2307.03172.

\bibitem[{Meta(2024)}]{meta2024llama33}
Meta. 2024.
\newblock \href {https://huggingface.co/meta-llama/Llama-3.3-70B-Instruct} {Llama 3.3 70b instruct}.
\newblock Hugging Face.

\bibitem[{OpenAI(2024{\natexlab{a}})}]{openai2024gpt4technicalreport}
OpenAI. 2024{\natexlab{a}}.
\newblock \href {https://arxiv.org/abs/2303.08774} {Gpt-4 technical report}.
\newblock \emph{Preprint}, arXiv:2303.08774.

\bibitem[{OpenAI(2024{\natexlab{b}})}]{openai_gpt41_2024}
OpenAI. 2024{\natexlab{b}}.
\newblock \href {https://openai.com/index/gpt-4-1/} {Introducing gpt-4.1 in the api}.
\newblock Accessed: 2025-05-17.

\bibitem[{OpenAI(2024{\natexlab{c}})}]{openai_o1_2024}
OpenAI. 2024{\natexlab{c}}.
\newblock Introducing openai o1.
\newblock \url{https://openai.com/o1/}.
\newblock Accessed: 2025-05-16.

\bibitem[{Pang et~al.(2022)Pang, Parrish, Joshi, Nangia, Phang, Chen, Padmakumar, Ma, Thompson, He, and Bowman}]{pang2022qualityquestionansweringlong}
Richard~Yuanzhe Pang, Alicia Parrish, Nitish Joshi, Nikita Nangia, Jason Phang, Angelica Chen, Vishakh Padmakumar, Johnny Ma, Jana Thompson, He~He, and Samuel~R. Bowman. 2022.
\newblock \href {https://arxiv.org/abs/2112.08608} {Quality: Question answering with long input texts, yes!}
\newblock \emph{Preprint}, arXiv:2112.08608.

\bibitem[{Research(2017)}]{faiss}
Facebook Research. 2017.
\newblock Faiss: A library for efficient similarity search and clustering of dense vectors.
\newblock \url{https://github.com/facebookresearch/faiss}.

\bibitem[{Trivedi et~al.(2022)Trivedi, Balasubramanian, Khot, and Sabharwal}]{trivedi2022musiquemultihopquestionssinglehop}
Harsh Trivedi, Niranjan Balasubramanian, Tushar Khot, and Ashish Sabharwal. 2022.
\newblock \href {https://arxiv.org/abs/2108.00573} {Musique: Multihop questions via single-hop question composition}.
\newblock \emph{Preprint}, arXiv:2108.00573.

\bibitem[{Wang et~al.(2024{\natexlab{a}})Wang, Ning, Pan, Wu, Guo, Deng, Bao, Hu, Zhang, Wang, and Zhang}]{wang2024novelqabenchmarkingquestionanswering}
Cunxiang Wang, Ruoxi Ning, Boqi Pan, Tonghui Wu, Qipeng Guo, Cheng Deng, Guangsheng Bao, Xiangkun Hu, Zheng Zhang, Qian Wang, and Yue Zhang. 2024{\natexlab{a}}.
\newblock \href {https://arxiv.org/abs/2403.12766} {Novelqa: Benchmarking question answering on documents exceeding 200k tokens}.
\newblock \emph{Preprint}, arXiv:2403.12766.

\bibitem[{Wang et~al.(2024{\natexlab{b}})Wang, Chen, Fu, Liao, Zhang, Wu, Yu, Xu, Zhang, Luo, Li, Yang, Huang, and Li}]{wang2024leavedocumentbehindbenchmarking}
Minzheng Wang, Longze Chen, Cheng Fu, Shengyi Liao, Xinghua Zhang, Bingli Wu, Haiyang Yu, Nan Xu, Lei Zhang, Run Luo, Yunshui Li, Min Yang, Fei Huang, and Yongbin Li. 2024{\natexlab{b}}.
\newblock \href {https://arxiv.org/abs/2406.17419} {Leave no document behind: Benchmarking long-context llms with extended multi-doc qa}.
\newblock \emph{Preprint}, arXiv:2406.17419.

\bibitem[{Welbl et~al.(2018)Welbl, Stenetorp, and Riedel}]{welbl2018constructingdatasetsmultihopreading}
Johannes Welbl, Pontus Stenetorp, and Sebastian Riedel. 2018.
\newblock \href {https://arxiv.org/abs/1710.06481} {Constructing datasets for multi-hop reading comprehension across documents}.
\newblock \emph{Preprint}, arXiv:1710.06481.

\bibitem[{Wu et~al.(2022)Wu, Rabe, Hutchins, and Szegedy}]{wu2022memorizingtransformers}
Yuhuai Wu, Markus~N. Rabe, DeLesley Hutchins, and Christian Szegedy. 2022.
\newblock \href {https://arxiv.org/abs/2203.08913} {Memorizing transformers}.
\newblock \emph{Preprint}, arXiv:2203.08913.

\bibitem[{Xu et~al.(2024)Xu, Ping, Wu, McAfee, Zhu, Liu, Subramanian, Bakhturina, Shoeybi, and Catanzaro}]{xu2024retrievalmeetslongcontext}
Peng Xu, Wei Ping, Xianchao Wu, Lawrence McAfee, Chen Zhu, Zihan Liu, Sandeep Subramanian, Evelina Bakhturina, Mohammad Shoeybi, and Bryan Catanzaro. 2024.
\newblock \href {https://arxiv.org/abs/2310.03025} {Retrieval meets long context large language models}.
\newblock \emph{Preprint}, arXiv:2310.03025.

\bibitem[{Yang et~al.(2018)Yang, Qi, Zhang, Bengio, Cohen, Salakhutdinov, and Manning}]{yang2018hotpotqadatasetdiverseexplainable}
Zhilin Yang, Peng Qi, Saizheng Zhang, Yoshua Bengio, William~W. Cohen, Ruslan Salakhutdinov, and Christopher~D. Manning. 2018.
\newblock \href {https://arxiv.org/abs/1809.09600} {Hotpotqa: A dataset for diverse, explainable multi-hop question answering}.
\newblock \emph{Preprint}, arXiv:1809.09600.

\bibitem[{Yuan et~al.(2024)Yuan, Ning, Zhou, Yang, Li, Zhuang, Tan, Yao, Lin, Li, Dai, Yan, and Wang}]{yuan2024lvevalbalancedlongcontextbenchmark}
Tao Yuan, Xuefei Ning, Dong Zhou, Zhijie Yang, Shiyao Li, Minghui Zhuang, Zheyue Tan, Zhuyu Yao, Dahua Lin, Boxun Li, Guohao Dai, Shengen Yan, and Yu~Wang. 2024.
\newblock \href {https://arxiv.org/abs/2402.05136} {Lv-eval: A balanced long-context benchmark with 5 length levels up to 256k}.
\newblock \emph{Preprint}, arXiv:2402.05136.

\bibitem[{Zaheer et~al.(2021)Zaheer, Guruganesh, Dubey, Ainslie, Alberti, Ontanon, Pham, Ravula, Wang, Yang, and Ahmed}]{zaheer2021bigbirdtransformerslonger}
Manzil Zaheer, Guru Guruganesh, Avinava Dubey, Joshua Ainslie, Chris Alberti, Santiago Ontanon, Philip Pham, Anirudh Ravula, Qifan Wang, Li~Yang, and Amr Ahmed. 2021.
\newblock \href {https://arxiv.org/abs/2007.14062} {Big bird: Transformers for longer sequences}.
\newblock \emph{Preprint}, arXiv:2007.14062.

\bibitem[{Zhang et~al.(2024)Zhang, Li, Liu, yang, Liu, Chen, Luo, and Yang}]{zhang2024marathonracerealmlong}
Lei Zhang, Yunshui Li, Ziqiang Liu, Jiaxi yang, Junhao Liu, Longze Chen, Run Luo, and Min Yang. 2024.
\newblock \href {https://arxiv.org/abs/2312.09542} {Marathon: A race through the realm of long context with large language models}.
\newblock \emph{Preprint}, arXiv:2312.09542.

\bibitem[{Zhu et~al.(2024)Zhu, Hwang, Dugan, and Callison-Burch}]{zhu2024fanoutqamultihopmultidocumentquestion}
Andrew Zhu, Alyssa Hwang, Liam Dugan, and Chris Callison-Burch. 2024.
\newblock \href {https://arxiv.org/abs/2402.14116} {Fanoutqa: A multi-hop, multi-document question answering benchmark for large language models}.
\newblock \emph{Preprint}, arXiv:2402.14116.

\end{thebibliography}

\clearpage
\onecolumn
\appendix

\clearpage


\section{Breakdown Visualizations of Model Accuracy Trends}
\label{sec:appendix-breakdowns}

\begin{figure}[H]
  \centering
  \includegraphics[width=\textwidth]{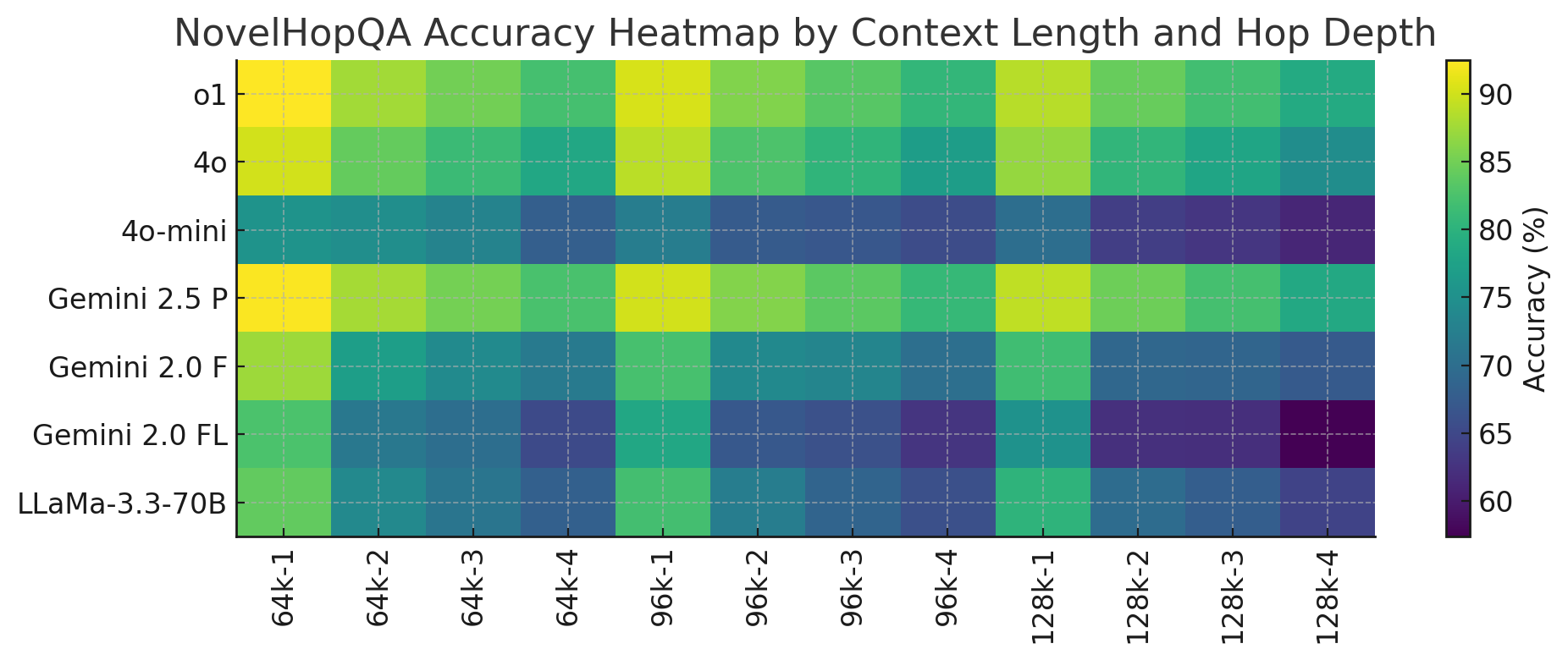}
  \caption{Accuracy (\%) on \textbf{\methodname} across context lengths and hop depths \( H \in \{1, 2, 3, 4\} \). This heatmap shows how model accuracy declines as both narrative length and multi-hop reasoning depth increase.}
  \label{fig:heatmap}
\end{figure}

To complement the heatmap, we include detailed line plots illustrating model-specific trends across each axis independently:

\begin{figure}[H]
  \centering
  \includegraphics[width=0.49\textwidth]{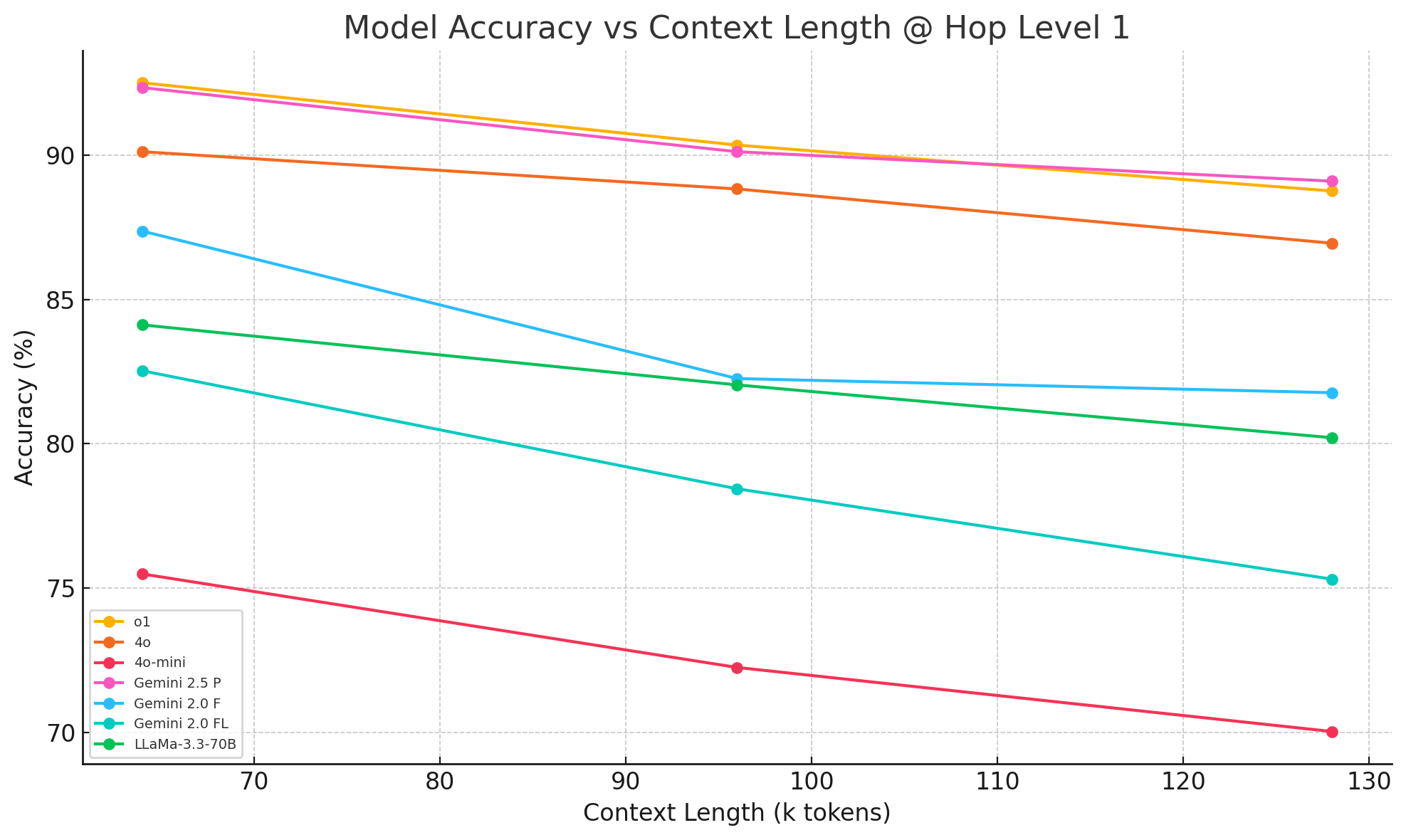}
  \includegraphics[width=0.49\textwidth]{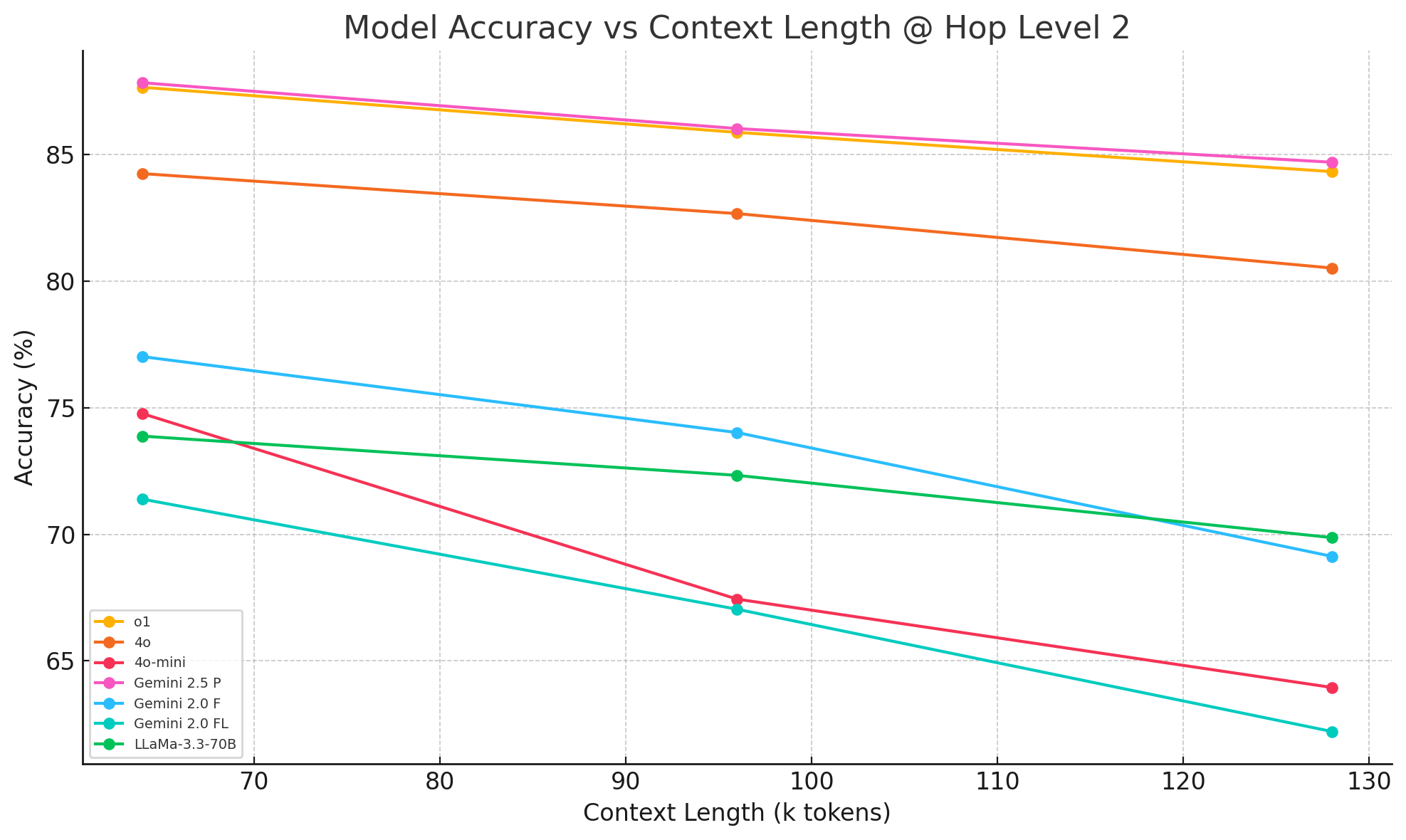}
  \includegraphics[width=0.49\textwidth]{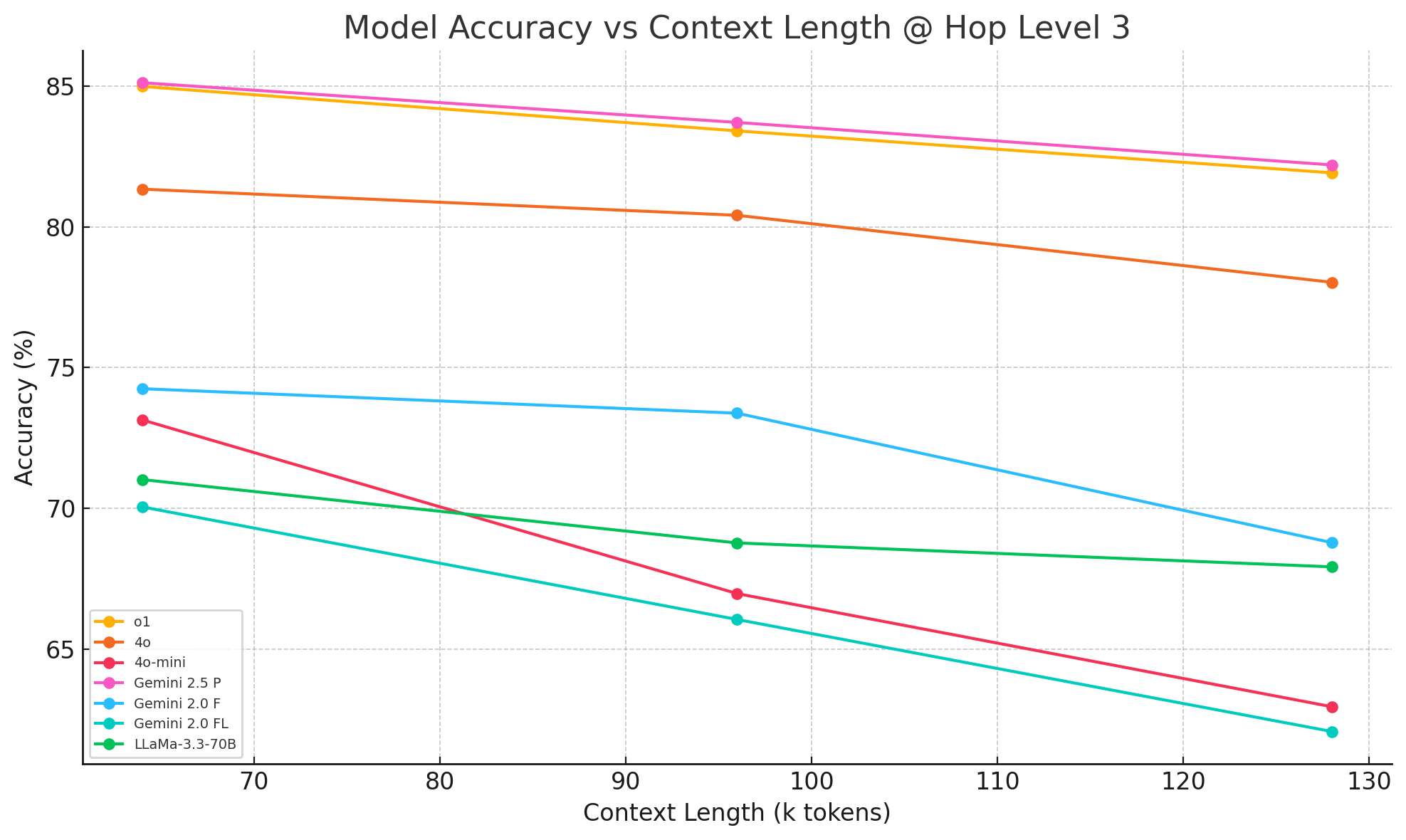}
  \includegraphics[width=0.49\textwidth]{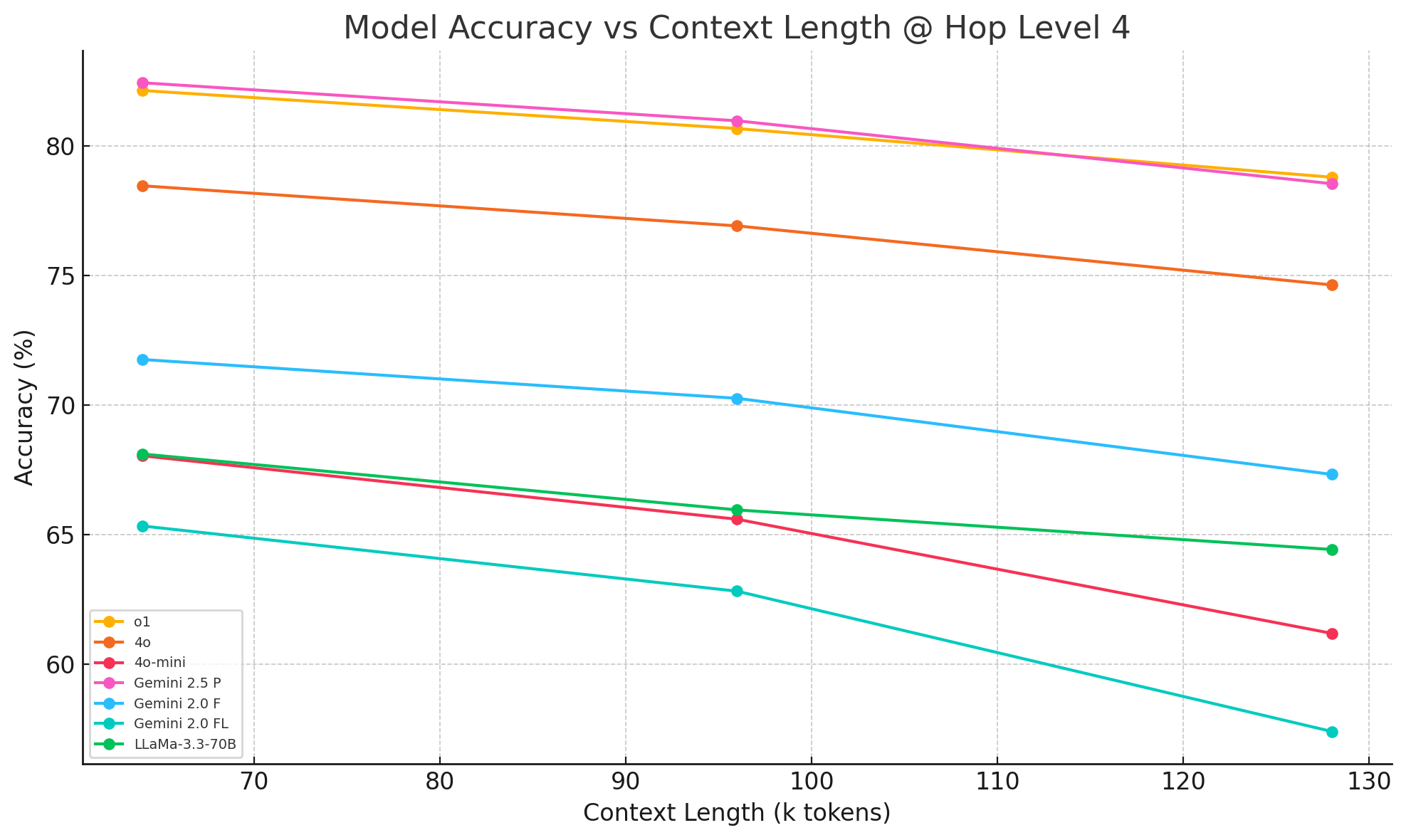}
  \caption{Model performance across context lengths for each hop level \( H = 1, 2, 3, 4 \). These plots isolate the effect of longer narratives on accuracy.}
  \label{fig:line-context}
\end{figure}

\vspace{1em}

\begin{figure}[H]
  \centering
  \includegraphics[width=0.49\textwidth]{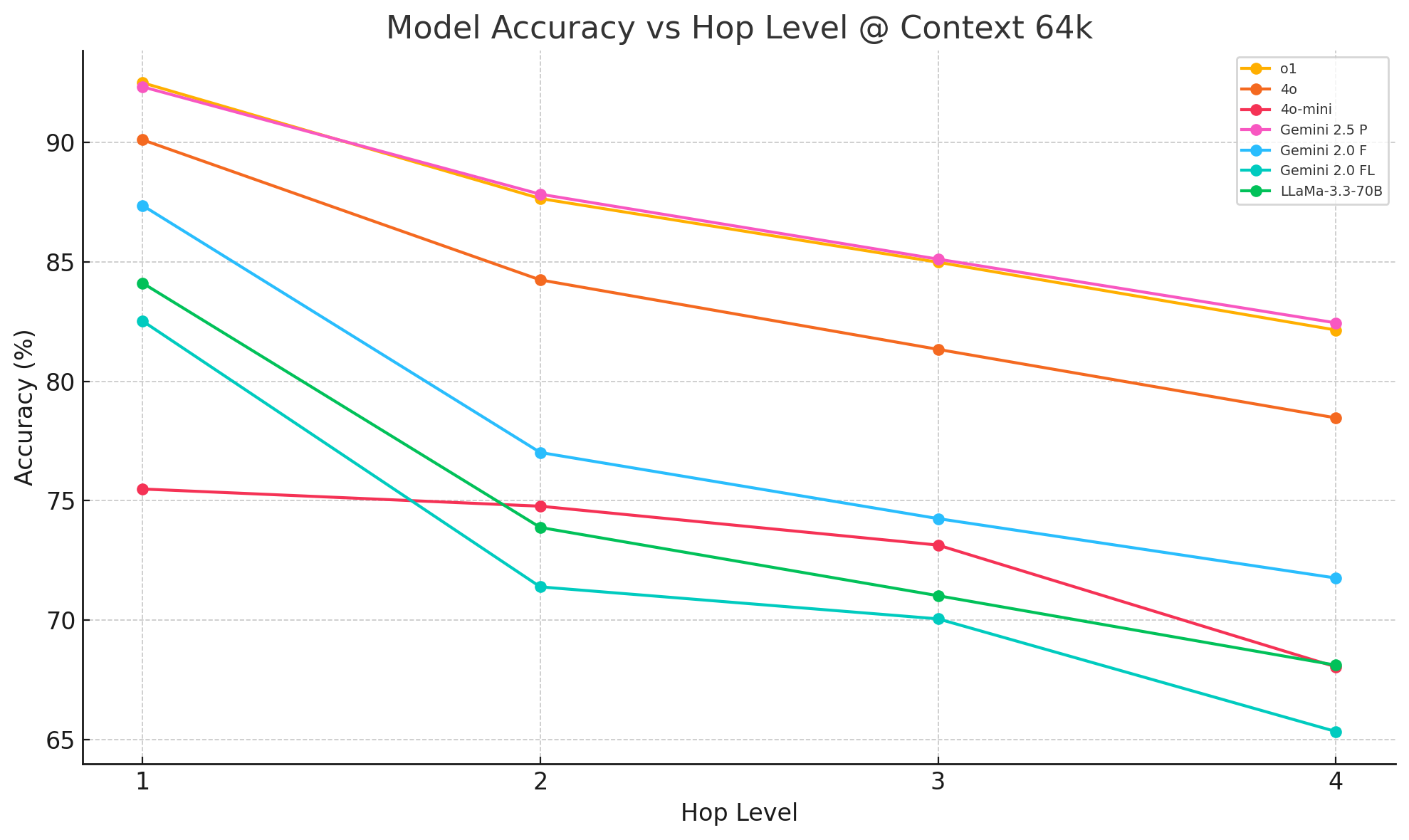}
  \includegraphics[width=0.49\textwidth]{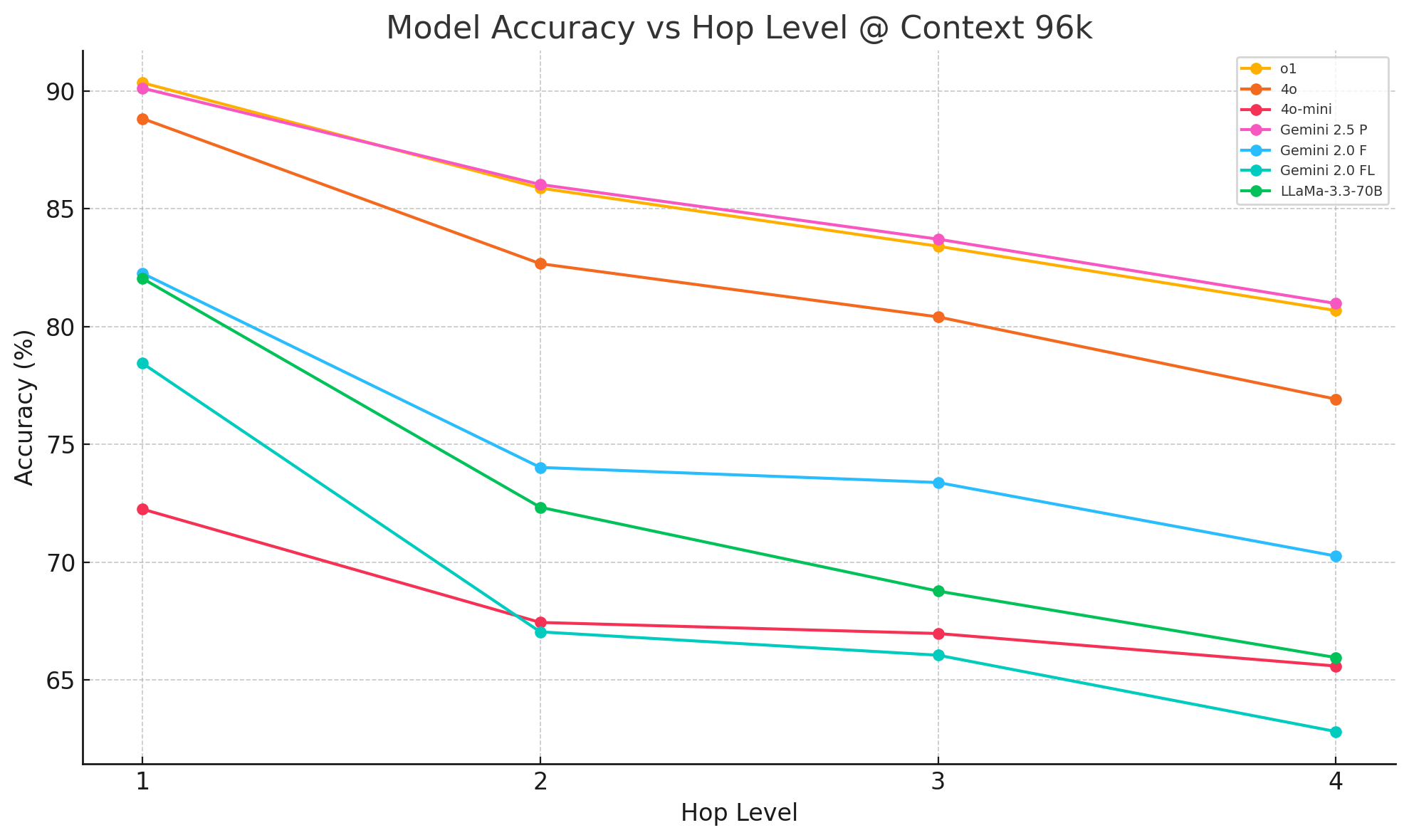}
  \includegraphics[width=0.49\textwidth]{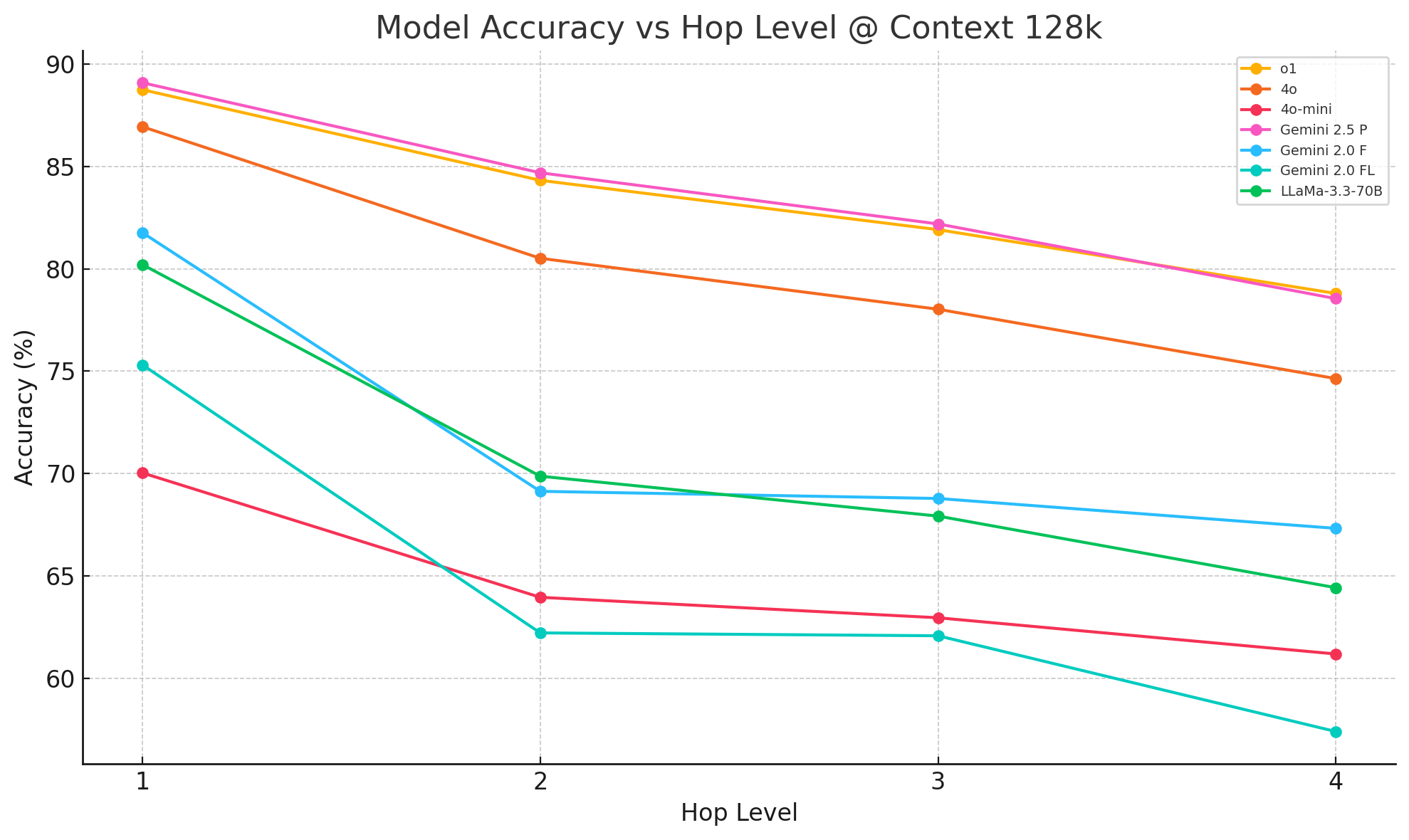}
  \caption{Model performance across hop levels for each context length (64k, 96k, 128k). These plots isolate the effect of deeper reasoning on accuracy.}
  \label{fig:line-hop}
\end{figure}

\section{Dataset Statistics by Hop Level}
\label{sec:appendix-hopstats}

\begin{table}[H]
  \centering
  \fontsize{11pt}{13pt}\selectfont
  \setlength{\tabcolsep}{10pt}
  \begin{tabular}{c|c|c|c}
    \toprule
    \textbf{Hop Level} & \textbf{Count} & \textbf{Avg. Context Tokens} & \textbf{Avg. Answer Length} \\
    \midrule
    1-Hop & 1000 & 191.92 & 4.64 \\
    2-Hop & 1000 & 451.46 & 6.99 \\
    3-Hop & 1000 & 691.85 & 9.59 \\
    4-Hop & 1000 & 916.82 & 10.79 \\
    \bottomrule
  \end{tabular}
  \caption{Dataset statistics across hop levels. Each row reports the number of QA pairs, the average context length in tokens, and the average answer length in words.}
  \label{tab:appendix-hop-stats}
\end{table}

\vspace{1em}
\subsection{Filtered Dataset Size After Golden-Context Evaluation}
\label{sec:appendix-filtered-dataset}

\begin{table}[H]
  \centering
  \fontsize{11pt}{13pt}\selectfont
  \setlength{\tabcolsep}{10pt}
  \begin{tabular}{c|c|c}
    \toprule
    \textbf{Hop Level} & \textbf{\# Removed} & \textbf{New Total} \\
    \midrule
    1-Hop & 37 & 963 \\
    2-Hop & 39 & 961 \\
    3-Hop & 40 & 960 \\
    4-Hop & 42 & 958 \\
    \bottomrule
  \end{tabular}
  \caption{Number of questions removed per hop after Golden-context filtering.}
  \label{tab:filtering-stats}
\end{table}

\vspace{-0.5em}

\section{Full Human Evaluation Table}
\label{sec:appendix-human-eval}

\begin{table}[H]
\centering
\small
\setlength{\tabcolsep}{8pt}
\begin{tabular}{c|cc|cc|cc|cc}
\toprule
\textbf{Validator} 
& \multicolumn{2}{c|}{\textbf{$H = 1$}} 
& \multicolumn{2}{c|}{\textbf{$H = 2$}} 
& \multicolumn{2}{c|}{\textbf{$H = 3$}} 
& \multicolumn{2}{c}{\textbf{$H = 4$}} \\
& \textbf{Align} & \textbf{Hop Match} 
& \textbf{Align} & \textbf{Hop Match} 
& \textbf{Align} & \textbf{Hop Match} 
& \textbf{Align} & \textbf{Hop Match} \\
\midrule
\textbf{Validator 1}  & 6.71 & 96.2 & 6.52 & 94.0 & 6.69 & 95.1 & 6.57 & 96.5 \\
\textbf{Validator 2}  & 6.66 & 97.1 & 6.43 & 95.3 & 6.55 & 93.6 & 6.64 & 94.9 \\
\textbf{Validator 3}  & 6.79 & 95.8 & 6.68 & 96.7 & 6.42 & 94.4 & 6.71 & 93.8 \\
\textbf{Validator 4}  & 6.60 & 94.7 & 6.57 & 93.9 & 6.61 & 95.2 & 6.45 & 96.1 \\
\textbf{Validator 5}  & 6.70 & 95.3 & 6.61 & 96.5 & 6.58 & 94.8 & 6.73 & 95.7 \\
\textbf{Validator 6}  & 6.58 & 96.9 & 6.65 & 95.2 & 6.66 & 96.6 & 6.52 & 94.5 \\
\textbf{Validator 7}  & 6.63 & 96.1 & 6.50 & 94.4 & 6.70 & 95.5 & 6.59 & 93.7 \\
\textbf{Validator 8}  & 6.74 & 95.0 & 6.56 & 93.6 & 6.47 & 94.3 & 6.65 & 96.8 \\
\textbf{Validator 9}  & 6.69 & 97.2 & 6.67 & 94.8 & 6.53 & 96.0 & 6.68 & 95.4 \\
\textbf{Validator 10} & 6.77 & 94.5 & 6.62 & 95.6 & 6.60 & 93.9 & 6.54 & 94.2 \\
\midrule
\textbf{Average}      
& \textbf{6.69} & \textbf{95.9} 
& \textbf{6.58} & \textbf{94.9} 
& \textbf{6.58} & \textbf{94.9} 
& \textbf{6.57} & \textbf{95.2} \\
\bottomrule
\end{tabular}
\caption{Full human validation scores across hop depths \( H \in \{1, 2, 3, 4\} \). “Alignment” is the average Likert rating (1–7); “Hop Match” is the percentage of responses judged to require exactly \( H \) reasoning steps.}
\label{tab:human_validators_scores}
\end{table}

\newpage
\clearpage

\section{Failure Mode Analysis}
\label{sec:failure_modes}

We isolate four clear-cut reasoning failures in questions, each demonstrated with an example. In every case, the gold answer provides exactly the required information from all reasoning steps, while the model answer either stops too early, confuses entities, omits part of the evidence, or drifts onto irrelevant details.

\subsection{1. Missing Final-Hop Integration}
Robust multi-hop reasoning requires chaining evidence through each of the four hops to reach a final conclusion. Here, the model successfully identifies the first three clues but then fails to incorporate the decisive testimony in hop 4, effectively truncating its reasoning chain. This indicates a breakdown in integrating the last piece of critical information.

\begin{table}[H]
\centering
\scriptsize
\setlength{\tabcolsep}{6pt}
\begin{tabular}{p{1cm}p{4.5cm}p{3.5cm}p{3.5cm}}
\toprule
\textbf{Hop} & \textbf{Question} & \textbf{Model Answer} & \textbf{Gold Answer} \\
\midrule
4 & After the council drafted a forged decree, encoded hidden warnings, left a fingerprint in the archives, and then overheard a sentry’s words, which testimony finally confirmed their betrayal?  
& The torn decree, the coded warnings, and the fingerprint.  
& The torn decree, the coded warnings, the fingerprint, and the sentry’s confession. \\
\bottomrule
\end{tabular}
\caption{The model omits the sentry’s confession in hop 4, showing it missed the final integration step.}
\end{table}

This example highlights how the model’s reasoning chain halts prematurely at hop 3, failing to incorporate the final piece of evidence that completes the inference.

\subsection{2. Entity Confusion / Coreference Errors}
Accurate multi-hop reasoning depends on consistently tracking entities across all hops. Ambiguous references or similar names can cause the model to substitute one entity for another in the final step. This reflects a coreference resolution failure that breaks the integrity of the entire reasoning chain.

\begin{table}[H]
\centering
\scriptsize
\setlength{\tabcolsep}{6pt}
\begin{tabular}{p{1cm}p{4.5cm}p{3.5cm}p{3.5cm}}
\toprule
\textbf{Hop} & \textbf{Question} & \textbf{Model Answer} & \textbf{Gold Answer} \\
\midrule
4 & After the knights gathered at dawn, rode through the Darkwood, crossed the Silver River, and repaired the collapsed causeway, which knight secured the bridge?  
& Sir Percival.  
& Sir Galahad. \\
\bottomrule
\end{tabular}
\caption{The model confuses Sir Galahad with Sir Percival in the final hop, misattributing the action.}
\end{table}

Here, a coreference error causes the model to swap one knight for another, illustrating how entity confusion derails multi-hop inference.

\subsection{3. Incomplete Evidence Combination}
Multi-hop questions demand synthesizing evidence from each of the four hops into a complete answer. A common failure is when the model extracts only a subset of the required evidences, indicating it did not fully aggregate all reasoning steps. This partial synthesis leaves out critical information.

\begin{table}[H]
\centering
\scriptsize
\setlength{\tabcolsep}{6pt}
\begin{tabular}{p{1cm}p{4.5cm}p{3.5cm}p{3.5cm}}
\toprule
\textbf{Hop} & \textbf{Question} & \textbf{Model Answer} & \textbf{Gold Answer} \\
\midrule
4 & When the telescope’s mirror cracked in the storm, its power supply surged, the control panel flickered, and temperatures spiked, what four malfunctions did the team record?  
& The cracked mirror, the power surge, and the flickering panel.  
& The cracked mirror, the power surge, the flickering panel, and the overheating coil. \\
\bottomrule
\end{tabular}
\caption{The model lists only hops 1–3 and omits the overheating coil from hop 4, showing incomplete evidence combination.}
\end{table}

This case demonstrates that the model gathers early clues but fails to include the final malfunction, indicating incomplete aggregation of all four pieces of evidence.

\subsection{4. Contextual Drift}
Sustained multi-hop reasoning requires maintaining focus on the relevant narrative thread. Over multiple hops, the model can drift back to an earlier, irrelevant detail, mistakenly including it instead of the true final clue. This reflects a failure to preserve contextual focus.

\begin{table}[H]
\centering
\scriptsize
\setlength{\tabcolsep}{6pt}
\begin{tabular}{p{1cm}p{4.5cm}p{3.5cm}p{3.5cm}}
\toprule
\textbf{Hop} & \textbf{Question} & \textbf{Model Answer} & \textbf{Gold Answer} \\
\midrule
4 & After the river swelled, the frogs fell silent, the oil lanterns sputtered, and compasses spun wildly, which four omens did villagers cite as signs of the flood?  
& The crimson sunset, the rising water, the silent frogs, and the spinning compass.  
& The rising water, the silent frogs, the sputtering lanterns, and the spinning compass. \\
\bottomrule
\end{tabular}
\caption{The model reintroduces “crimson sunset” from an early pass, demonstrating drift away from the four correct omens.}
\end{table}

This example shows how the model’s attention drifts to a decorative detail from hop 1, instead of preserving focus on the four true flood omens.

\newpage
\clearpage

\section{Irrelevant and No Context Evaluation}
\label{sec:appendix-fake-context}

To evaluate whether models genuinely rely on the narrative context provided in \textbf{\methodname}, we conduct an ablation study using two control conditions: \textbf{irrelevant context} and \textbf{no context}. This analysis verifies that model accuracy is not attributable to memorization or dataset leakage.

\vspace{0.5em}
\noindent
\textbf{Irrelevant Context.} For each question, we prompted the model with unrelated context. The paragraph has no semantic or lexical relationship to the QA pair. The irrelevant context used is shown in Appendix Table~\ref{tab:fake-context-example}.

\vspace{0.3em}
\noindent
\textbf{No Context.} The model is given only the question and no surrounding passage. This isolates performance that arises solely from model priors or memorized facts.

\vspace{0.5em}
\noindent
\textbf{Experimental Setup.} Each model was evaluated on 800 examples—100 random questions from each of four datasets, under both irrelevant and no context conditions. All responses were graded by GPT‑4.1 \cite{openai_gpt41_2024} using CoT prompting for consistency.

\begin{table*}[h]
\centering
\footnotesize
\renewcommand{\arraystretch}{1.25}
\setlength{\tabcolsep}{10pt}
\begin{tabular}{@{}llcccc@{}}
\toprule
\textbf{Model} & \textbf{Condition} & \textbf{1-hop} & \textbf{2-hop} & \textbf{3-hop} & \textbf{4-hop} \\
\midrule

\multirow{2}{*}{\texttt{Gemini 2.0 Flash Lite}} 
  & Irrelevant context & 4\% (4/100) & 3\% (3/100) & 1\% (1/100) & 1\% (1/100) \\
  & No context         & 4\% (4/100) & 3\% (3/100) & 1\% (1/100) & 1\% (1/100) \\

\midrule
\multirow{2}{*}{\texttt{GPT-4o Mini}} 
  & Irrelevant context & 5\% (5/100) & 4\% (4/100) & 1\% (1/100) & 1\% (1/100) \\
  & No context         & 4\% (4/100) & 3\% (3/100) & 1\% (1/100) & 1\% (1/100) \\

\midrule
\multirow{2}{*}{\texttt{Gemini 2.0 Flash}} 
  & Irrelevant context & 6\% (6/100) & 5\% (5/100) & 1\% (1/100) & 1\% (1/100) \\
  & No context         & 5\% (5/100) & 4\% (4/100) & 1\% (1/100) & 1\% (1/100) \\

\midrule
\multirow{2}{*}{\texttt{GPT-4o}} 
  & Irrelevant context & 6\% (6/100) & 5\% (5/100) & 2\% (2/100) & 1\% (1/100) \\
  & No context         & 6\% (6/100) & 5\% (5/100) & 1\% (1/100) & 1\% (1/100) \\

\midrule
\multirow{2}{*}{\texttt{o1}} 
  & Irrelevant context & 6\% (6/100) & 5\% (5/100) & 2\% (2/100) & 1\% (1/100) \\
  & No context         & 6\% (6/100) & 5\% (5/100) & 2\% (2/100) & 1\% (1/100) \\

\midrule
\multirow{2}{*}{\texttt{Gemini 2.5 Pro}} 
  & Irrelevant context & 7\% (7/100) & 6\% (6/100) & 2\% (2/100) & 1\% (1/100) \\
  & No context         & 7\% (7/100) & 5\% (5/100) & 2\% (2/100) & 1\% (1/100) \\

\midrule
\multirow{2}{*}{\texttt{LLaMA 3.3 70B Instruct}} 
  & Irrelevant context & 6\% (6/100) & 3\% (3/100) & 2\% (2/100) & 1\% (1/100) \\
  & No context         & 5\% (5/100) & 4\% (4/100) & 1\% (1/100) & 2\% (2/100) \\

\bottomrule
\end{tabular}
\caption{Accuracy (\%) on 100 randomly selected multi-hop questions under irrelevant and no context settings. Models perform poorly across all hops, demonstrating that answers cannot be derived without relevant narrative input.}
\end{table*}

\vspace{-1em}

\begin{table}[H]
\centering
\begin{minipage}{\linewidth}
\begin{tcolorbox}[
    colback=gray!4!white,
    colframe=gray!60!black,
    title=Irrelevant Context Example (\emph{The Secret Garden}),
    fonttitle=\bfseries,
    width=\linewidth,
    boxsep=3pt,
    left=4pt,
    right=4pt,
    top=2pt,
    bottom=2pt]
\scriptsize
\textbf{Context Source:}

\begin{enumerate}[nosep,leftmargin=1.2em]
  \item \textbf{Paragraph 1.} \\
  It was the sweetest, most mysterious-looking place any one could imagine. The high walls which shut it in were covered with the leafless stems of climbing roses which were so thick that they were matted together. Mary Lennox knew they were roses because she had seen a great many roses in India. All the ground was covered with grass of a wintry brown, and out of it grew clumps of bushes which were surely rose-bushes if they were anything. There were numbers of standard roses which had so spread their branches that they were like little trees. There were other trees in the garden, and one of the things which made the place look strangest and loveliest was that climbing roses had run all over them and swung down long tendrils which made light swaying curtains.
  
  \item \textbf{Paragraph 2.} \\
  And here and there among the grass were narcissus bulbs beginning to sprout and uncurl their narrow green leaves. She thought they seemed to be stretching out their arms to see how warm the sun was. She went from one part of the garden to another. She found many more of the sprouting pale green points and she found others which were white crocuses and snowdrops, because the green spikes had burst through their sheaths and showed white. She remembered what Ben Weatherstaff had said about the “snowdrops by the thousands,” and about bulbs spreading and making new ones. “These had been left to themselves for ten years,” perhaps, and they had spread like the snowdrops into thousands.
\end{enumerate}
\end{tcolorbox}
\end{minipage}
\caption{The “irrelevant context” passage used during ablation. This excerpt, unrelated to any QA pair, was paired with a question to test whether models output plausible answers.}
\label{tab:fake-context-example}
\end{table}

\noindent
\paragraph{Interpretation.} This experiment validates the integrity of \textbf{\methodname} by confirming that models are not simply memorizing QA pairs seen during pretraining. Accuracy remains near-zero when relevant context is removed, demonstrating that our questions are novel and context-dependent. These findings strengthen confidence that model performance on \textbf{\methodname} reflects actual reading comprehension and not artifact exploitation or memorization.

\section{RAG Evaluations}
\label{sec:rag-evals}

\paragraph{Pipeline.}  
We divide each novel into non-overlapping chunks of \textbf{350 tokens}, which are embedded using the \textbf{bge-large-en} encoder~\cite{bge-large-en}. During inference, we use the \textbf{Facebook FAISS} retrieval model~\cite{faiss} to select the top $k{=}7$ most relevant chunks based on inner-product similarity.

At inference time, we follow this process:

\begin{enumerate}[nosep,leftmargin=*]
  \item Encode the input question using bge-large-en.
  \item Use the Facebook FAISS retrieval model to find the \textbf{top $k{=}7$ most relevant chunks} via inner-product similarity search.
  \item Concatenate the retrieved chunks in their original order and prepend the question to form a context of roughly 2.5k tokens.
  \item Pass this context to each of the seven models.
\end{enumerate}

\paragraph{Why a RAG setting?}
The retrieved context mimics a real-world system in which only \emph{relevant snippets}—not the full 64k–128k window—are available to the generator.  
We expect lower accuracy because (i) retrieval can miss one or more hop paragraphs and (ii) evidence may be partial or out of order.

\paragraph{Results.}
Table~\ref{tab:rag-results} reports accuracy on the same questions.  
As anticipated, scores cluster around 50 \%, roughly 25–35 points below the golden-context setting.

\begin{table}[H]
\centering
\scriptsize
\setlength{\tabcolsep}{10pt}
\begin{tabular}{c|ccccccc|c}
\toprule
\textbf{Hop} & \textbf{o1} & \textbf{4o} & \textbf{4o-mini} & \textbf{Gemini 2.5 P} & \textbf{Gemini 2.0 F} & \textbf{Gemini 2.0 FL} & \textbf{LLaMa-3.3-70B-Instruct} & \textbf{Avg.}\\
\midrule
1 & 62.43 & 60.87 & 49.18 & \textbf{63.14} & 55.03 & 43.92 & 48.23 & 54.69\\
2 & 56.03 & 54.78 & 46.27 & \textbf{57.36} & 49.86 & 39.29 & 44.69 & 49.75\\
3 & 52.13 & 50.37 & 43.78 & \textbf{54.04} & 46.47 & 40.68 & 42.93 & 47.20\\
4 & 48.35 & 47.05 & 40.97 & \textbf{50.31} & 43.63 & 36.57 & 39.83 & 43.82\\
\midrule
\textbf{Avg.} & 54.74 & 53.27 & 45.05 & \textbf{56.21} & 48.75 & 40.11 & 43.92 & 48.86\\
\bottomrule
\end{tabular}
\caption{Accuracy (\%) of RAG-augmented models. For each query, the retriever encoder processed the full novel to generate vector embeddings for all chunks, which were then used to retrieve the top $k{=}7$ most relevant chunks.}
\label{tab:rag-results}
\end{table}

\paragraph{Analysis.}
Despite feeding models the full retrieved context directly in the input prompt, \textbf{LLMs still underperform.}. This gap underscores the limitations of retrieval-augmented reasoning even when relevant evidence is available at inference time. Our error analysis identifies two dominant failure modes:

\begin{enumerate}[nosep,leftmargin=*]
    \item \textbf{Incomplete hop coverage} — For multi-hop questions, crucial context is often fragmented across the book. When even one necessary chunk is missing from the top-$k$ retrieved segments, models fail to complete the reasoning chain. This issue becomes more pronounced in 3- and 4-hop examples, where partial evidence leads to confidently incorrect answers.
    
    \item \textbf{Evidence misalignment} — Unlike the golden oracle where evidence is presented in coherent order, RAG-retrieved chunks may be out of narrative sequence or lack discourse continuity. This can confuse even strong models, leading to misinterpretation of character arcs, event timelines, or causal links.
\end{enumerate}
\vspace{1.5em}
\noindent{These failure modes point to the brittleness of current RAG pipelines when applied to deep reasoning tasks over long texts. They support our broader conclusion: \textbf{retrieval helps, but cannot replace full-context comprehension in complex multi-hop reasoning}. Addressing this limitation may require more sophisticated retrieval strategies—such as iterative chunk expansion, retrieval-conditioned generation, or train-time exposure to fragmented narratives.}

\newpage

\subsection{Visual Breakdown of RAG Accuracy}

\begin{figure}[H]
  \centering
  \includegraphics[width=0.42\textwidth]{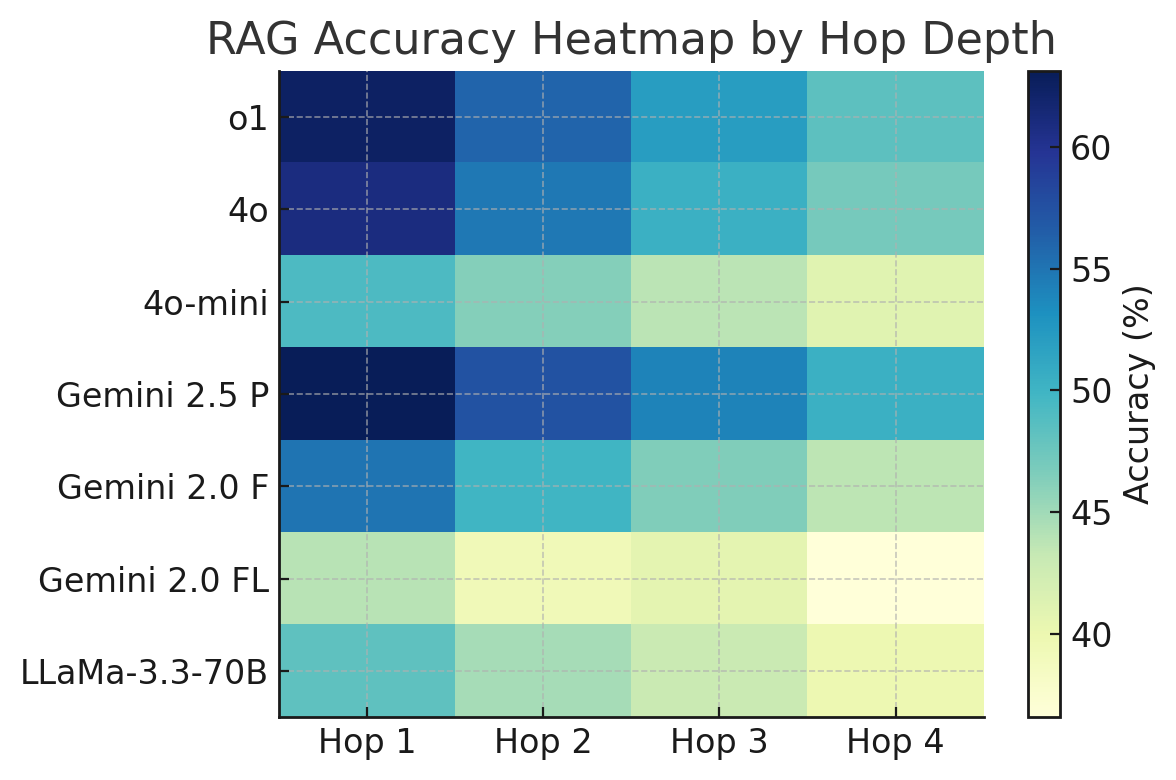}\hfill
  \includegraphics[width=0.50\textwidth]{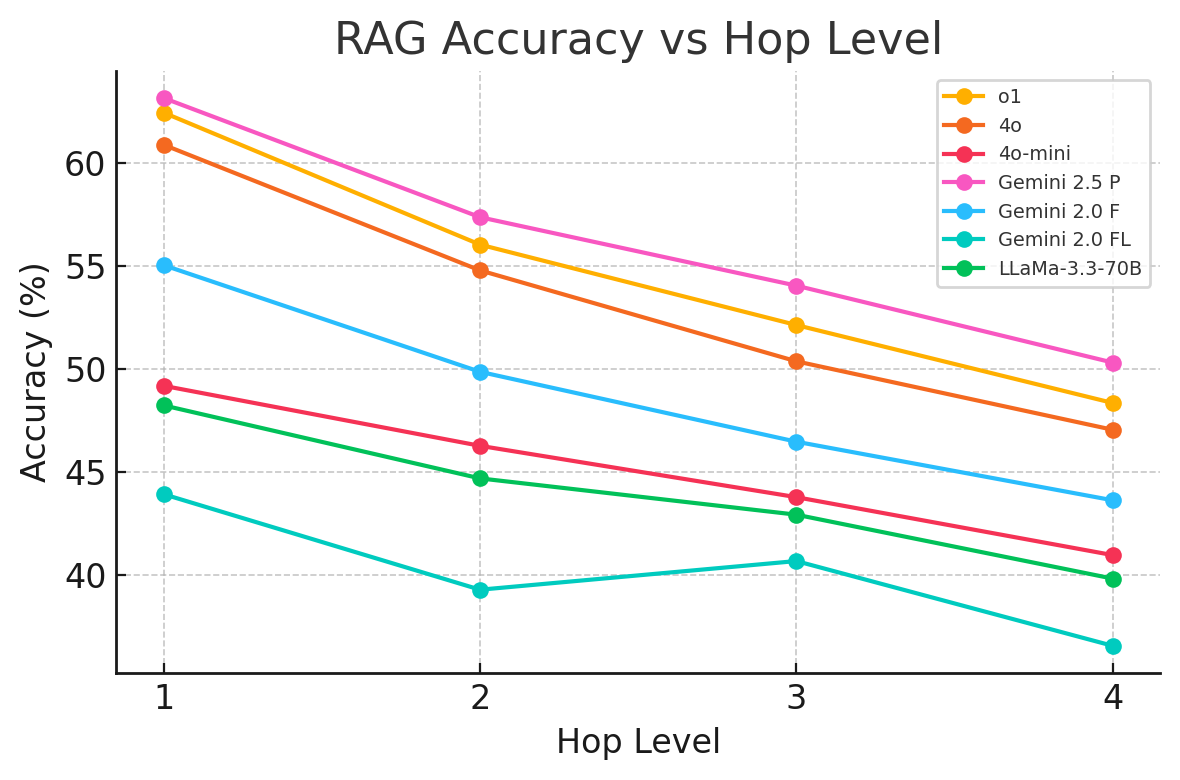}

  \caption{Left: Heatmap of RAG-augmented accuracy across hop depths for all models.  
  Right: Accuracy curves for each model as a function of hop level. Together, these plots show how performance degrades with deeper reasoning steps even when retrieval narrows the context.}
  \label{fig:rag-heat-line}
\end{figure}

\section{4-Hop QA Evolution Example}
\label{sec:example-four-hop}

\begin{table}[H]
\centering
\begin{minipage}{\linewidth}
\begin{tcolorbox}[
    colback=gray!4!white,
    colframe=gray!60!black,
    title=4-Hop QA Evolution Example,
    fonttitle=\bfseries,
    width=\linewidth,
    boxsep=3pt,
    left=4pt,
    right=4pt,
    top=2pt,
    bottom=2pt]
\scriptsize
\textbf{Context Source (Original Narrative):}

\begin{enumerate}[nosep,leftmargin=1.2em]
  \item \textbf{Paragraph 1 (Attic Discovery).}\\
        In the dusty attic, she uncovered a half-torn parchment depicting a faded map with a bold “X” at its center.
  \item \textbf{Paragraph 2 (Courtyard Statue).}\\
        The same “X” symbol was intricately carved into the base of the knight’s statue in the castle courtyard.
  \item \textbf{Paragraph 3 (Hidden Chamber).}\\
        A secret latch at the statue’s pedestal opened to reveal a small iron key etched with the image of a rising phoenix.
  \item \textbf{Paragraph 4 (Oak Tree Pedestal).}\\
        Beneath the ancient oak tree in the east garden, a stone pedestal bore a matching keyhole, sealed until the phoenix key was inserted.
\end{enumerate}

\vspace{0.6em}
\textbf{Questions Across Hops:}
\begin{itemize}[nosep,leftmargin=1.4em]
  \item \textbf{Hop 1:} What object did she find in the attic that launched her quest?  
  \item \textbf{Hop 2:} Based on the attic map’s “X” and the statue carving, which landmark did she identify to continue her search?  
  \item \textbf{Hop 3:} After locating the statue, what item did she retrieve from its hidden compartment to aid her quest?  
  \item \textbf{Hop 4:} How did she ultimately open the pedestal beneath the oak tree using the map, the statue clues, and the retrieved key?  
\end{itemize}

\vspace{0.3em}
\textbf{Final Answer (Hop 4):}\\
She first found the half-torn parchment map, then matched its “X” to the courtyard knight’s statue, retrieved the phoenix-etched iron key from the statue’s compartment, and finally inserted that key into the oak tree pedestal’s keyhole to open it.

\vspace{0.4em}
\textbf{Hop Reasoning Breakdown:}
\begin{itemize}[nosep,leftmargin=1.4em]
  \item \textbf{Hop 1 — Map Discovery}: Finds the parchment map in the attic.  
  \item \textbf{Hop 2 — Landmark Identification}: Uses the map’s “X” and statue carving to locate the courtyard statue.  
  \item \textbf{Hop 3 — Key Retrieval}: Opens the statue’s compartment and retrieves the phoenix-etched key.  
  \item \textbf{Hop 4 — Pedestal Unlock}: Uses the retrieved key with map/statue clues to open the oak tree pedestal.  
\end{itemize}
\end{tcolorbox}
\end{minipage}
\caption{4-hop QA example showing the step-wise evolution of context, question, and reasoning.}
\end{table}

\vspace{-.5em}
\section{Human Evaluation Form Example}
\label{sec:human-eval-form}

\begin{table}[H]   
\centering
\begin{minipage}{\linewidth}
\begin{tcolorbox}[
    colback=gray!4!white,
    colframe=gray!60!black,
    title=Human Evaluation Form (3-Hop),
    fonttitle=\bfseries,
    width=\linewidth,
    boxsep=3pt,
    left=4pt,
    right=4pt,
    top=2pt,
    bottom=2pt]
\footnotesize\ttfamily
\noindent
\textbf{Paragraph 1:}\\
Now, inclusive of the occasional wide intervals between the revolving outer circles, and inclusive of the spaces between the various pods in any one of those circles, the entire area at this juncture, embraced by the whole multitude, must have contained at least two or three square miles. […] Queequeg patted their foreheads; Starbuck scratched their backs with his lance; but fearful of the consequences, for the time refrained from darting it.

\vspace{0.5em}
\textbf{Paragraph 2:}\\
But not a bit daunted, Queequeg steered us manfully; now sheering off from this monster directly across our route in advance; now edging away from that, whose colossal flukes were suspended overhead, while all the time, Starbuck stood up in the bows, lance in hand, pricking out of our way whatever whales he could reach. […]

\vspace{0.5em}
\textbf{Paragraph 3:}\\
“I will have the first sight of the whale myself,”—he said. […] Then arranging his person in the basket, he gave the word for them to hoist him to his perch, \textbf{Starbuck} being the one who secured the rope at last; and afterwards stood near it. […]

\vspace{0.5em}
\textbf{Question:} Why was Starbuck—rather than Queequeg—responsible for securing Captain Ahab’s rope before Ahab was hoisted to his perch?

\vspace{0.6em}
\textbf{Is this a 3-hop question?} (circle one)\\
Yes \hspace{2em} No

\vspace{0.6em}
\textbf{Rate alignment on a 7-point Likert scale} (circle one):

1 – Completely unrelated, 2 – Mostly unrelated, 3 – Somewhat related, 4 – Moderately related, 5 – Strongly related, 6 – Very closely related, 7 – Perfectly aligned
\end{tcolorbox}
\end{minipage}
\caption{Example form used by validators to assess hop depth and contextual alignment.}
\end{table}

\FloatBarrier

\clearpage

\section{Prompt Templates}
\label{sec:appendix-prompts}

\begin{tcolorbox}[
    colback=gray!4!white,
    colframe=gray!60!black,
    title=Anchor Keyword Generation,
    fonttitle=\bfseries,
    width=\textwidth,
    boxsep=3pt,
    left=4pt,
    right=4pt,
    top=2pt,
    bottom=2pt,
    before skip=10pt,
    after skip=10pt
]
\ttfamily
You are a literary analysis expert. Based solely on the book title “\{book\_title\}”, list five main keywords central to its plot. Ensure each keyword is concise (one or two words) and appears at least 50 times.

Answer format:
\begin{verbatim}
<keyword_result>
keyword1;
keyword2;
keyword3;
keyword4;
keyword5
</keyword_result>
\end{verbatim}
\end{tcolorbox}
\captionsetup{type=figure}
\captionof{figure}{Prompt for extracting five high-frequency anchor keywords from a book title}

\begin{tcolorbox}[
    colback=gray!4!white,
    colframe=gray!60!black,
    title=Single Hop Generation,
    fonttitle=\bfseries,
    width=\textwidth,
    boxsep=3pt,
    left=4pt,
    right=4pt,
    top=2pt,
    bottom=2pt,
    before skip=10pt,
    after skip=10pt
]

\ttfamily
You are an expert question generator. Given the paragraph below, generate one challenging question that requires understanding of this paragraph. Provide a concise answer.

Output format:
\begin{verbatim}
<question>Your question here</question>
<answer>Your concise answer here</answer>
\end{verbatim}

Paragraph: \{paragraph\}
\end{tcolorbox}
\captionsetup{type=figure}
\captionof{figure}{Prompt for generating a single-hop question from one paragraph.}

\begin{tcolorbox}[
    colback=gray!4!white,
    colframe=gray!60!black,
    title=Extract Related Keyword,
    fonttitle=\bfseries,
    width=\textwidth,
    boxsep=3pt,
    left=4pt,
    right=4pt,
    top=2pt,
    bottom=2pt,
    before skip=10pt,
    after skip=10pt
]

\ttfamily
You are an expert at extracting related keywords. From the paragraph below, identify a keyword strongly related to its content but different from “\{current\_keyword\}”. Return only the new keyword.

Output format:
\begin{verbatim}
<keyword>NEW_KEYWORD</keyword>
\end{verbatim}

Paragraph: \{paragraph\}
\end{tcolorbox}
\captionsetup{type=figure}
\captionof{figure}{Prompt for extracting a related keyword at hop h.}

\begin{tcolorbox}[
    colback=gray!4!white,
    colframe=gray!60!black,
    title=Generate Final Multi-Hop Question,
    fonttitle=\bfseries,
    width=\textwidth,
    boxsep=3pt,
    left=4pt,
    right=4pt,
    top=2pt,
    bottom=2pt,
    before skip=10pt,
    after skip=10pt
]

\ttfamily
You are an expert multi‑hop question generator. Generate one question requiring integration across all provided paragraphs, and provide a concise answer.

Output format:
\begin{verbatim}
<question>Your multi‑hop question here</question>
<answer>Your concise answer here</answer>
\end{verbatim}

Context:
\{paragraph1\}\textbackslash n\textbackslash n
\{paragraph2\}\,...\textbackslash n\textbackslash n
\{paragraphH\}
\end{tcolorbox}
\captionsetup{type=figure}
\captionof{figure}{Prompt for generating the final multi-hop question over H paragraphs.}

\end{document}